\documentclass{article}
\usepackage{PRIMEarxiv}
\PassOptionsToPackage{square,numbers,sort}{natbib}

\usepackage{pgfplots}
\usetikzlibrary{pgfplots.groupplots}
\usetikzlibrary{positioning,shapes,arrows,patterns,fit,backgrounds,calc,svg.path,3d,decorations.text,shapes.arrows,spy}
\usepackage{placeins}
\usepackage[utf8]{inputenc} % allow utf-8 input
\usepackage[T1]{fontenc}    % use 8-bit T1 fonts
\usepackage{hyperref}       % hyperlinks
\usepackage{url}            % simple URL typesetting
\usepackage{booktabs}       % professional-quality tables
\usepackage{amsfonts}       % blackboard math symbols
\usepackage{nicefrac}       % compact symbols for 1/2, etc.
\usepackage{microtype}      % microtypography
\usepackage{xcolor}         % colors
\usepackage{amsmath,amssymb}
\usepackage{algorithm}
\usepackage{algpseudocode}
\DeclareMathOperator*{\argmax}{arg\,max}
\DeclareMathOperator*{\argmin}{arg\,min}
\title{Distribution Mismatch Correction for Improved Robustness in Deep Neural Networks}
\author{%
  Alexander Fuchs, Christian Knoll, Franz Pernkopf\\
  Signal Processing and Speech Communication Laboratory\\
  Graz University of Technology\\
  \texttt{fuchs@tugraz.at}
}

\begin{document}
\maketitle

\begin{abstract}
Deep neural networks rely heavily on normalization methods to improve their performance and learning behavior. 
Although normalization methods spurred the development of increasingly deep and efficient architectures, they also increase the vulnerability with respect to noise and input corruptions.
In most applications, however, noise is ubiquitous and diverse; this can often lead to complete failure of machine learning systems as they fail to cope with mismatches between the input distribution during training- and test-time. The most common normalization method, batch normalization, reduces the distribution shift during training but is agnostic to changes in the input distribution during test time. This makes batch normalization prone to performance degradation whenever noise is present during test-time. Sample-based normalization methods can correct linear transformations of the activation distribution but cannot mitigate changes in the distribution shape; this makes the network vulnerable to distribution changes that cannot be reflected in the normalization parameters. We propose an unsupervised non-parametric distribution correction method that adapts the activation distribution of each layer. This reduces the mismatch between the training and test-time distribution by minimizing the 1-D Wasserstein distance. In our experiments, we empirically show that the proposed method effectively reduces the impact of intense image corruptions and thus improves the classification performance without the need for retraining or fine-tuning the model.
\end{abstract}

\section{Introduction}\label{sec:introduction}
Early on, Neural Networks (NNs) have proven to excel at interpolating between training data points but to fail when extrapolating to regions not covered by the training data~\cite{158898,227294}. The lack of sufficiently large datasets, therefore, limited the application of NNs to tasks with well-known input distributions; this prevents any unpredictable behavior in the network that might stem from data with unknown distribution. More precisely -- given a model with parameters $\mathbf{\theta}$, output $\mathbf{y}$, input $\mathbf{x}$, and a static conditional probability $p(\mathbf{y}|\mathbf{x},\mathbf{\theta})$ -- the problem of evaluating samples  $\mathbf{x}$ from a different distribution than the training distribution is known as \emph{covariate shift}~\cite{shimodaira2000improving,sugiyama2007covariate}. 

In recent years, the rise of big data and data augmentation techniques have alleviated the problem of distribution shifts via increasing the number of samples from the input space~\cite{krizhevsky2012imagenet}. The problem of covariate shift, however, remained – albeit in a different form: when training Deep Neural Networks (DNNs), each parameter update causes a distribution shift for the next mini-batch in the consecutive layers, resulting in convergence problems for DNNs. In particular with DNNs the problem is further exacerbated by a large number of layers since the distribution shifts can occur internally (within the model) before every layer. Thus, the main impact of the covariate shift moved from test-time to training-time. 

The introduction of \emph{Batch Normalization} (BN) reduced such internal covariate shifts during training by matching the distribution of activations across batches and, in doing so, greatly improved the convergence of deep Convolutional Neural Networks (CNNs)~\cite{DBLP:journals/corr/IoffeS15}.  
This improvement has fueled the development of ever deeper and more capable architectures and -- by reducing the dependence on the weight initialization -- facilitated training networks in an end-to-end fashion. Therefore, normalization became an elemental part of all deep learning architectures.
But while BN reduces the problem of shifts in the activation distributions during training, it estimates the normalization parameters (i.e., the mean $\mu$ and variance $\sigma^2$) according to the expectation over all training samples and thus remains agnostic to changes in the input distribution during test-time.
This makes BN inherently vulnerable to such changes that are e.g. caused by image corruptions~\cite{Benz_2021_WACV} and forces the models to extrapolate to regions not covered during training. To improve the corruption robustness of DNNs, one can expand the training space by data augmentation. Although this improves robustness against specific types of corruption, it concurrently reduces robustness against other types of corruptions~\cite{pmlr-v97-gilmer19a,Benz_2021_WACV}. This highlights the importance of mitigating covariate shifts during test-time.

\emph{Group Normalization} (GN) and \emph{Filter Response Normalization} (FRN) have been proposed to overcome the batch size dependence of BN (caused by insufficient statistics for small batches). Both methods are more flexible than BN as they compute the normalization parameter over individual samples and are thus more robust against changes in the activation distributions. Moreover, they perform comparably to BN for most classification tasks~\cite{wu2018group,singh2020filter}. Nonetheless, BN achieves state-of-the-art results for most architectures and remains the most commonly used normalization method, while the more robust alternatives GN and FRN are scarcely used. One important limitation of all the above normalization methods is that they can only correct for linear distribution transformations (e.g. mean shift or variance scaling) but not for mismatches between the shape of the distribution.

We propose a non-parametric distribution correction method that utilizes the 1D-Wasserstein distance and reduces distribution mismatches of arbitrary form during test-time. This correction method can be combined with other normalization methods and corrects for changes in the distribution shape in an unsupervised setting without the need for retraining or fine-tuning of the models, in contrast to self-supervised methods which adapt at least some of the model parameters \cite{schneider2020improving,liang2020we,sun2020test,wang2021tent}.
Our proposed approach is an iterative procedure following an energy minimization scheme as used in image denoising~\cite{perona1990scale,buades2005review,rudin1992nonlinear,Zhu08anefficient}. It is agnostic to the specific type of noise and maps all noisy activations to the target distribution of each layer. 

The target distribution is constructed on the basis of the typical activation distribution during training and is represented by its Wasserstein barycenter.
We compare the target to the test-time distribution after each activation layer and -- if necessary -- calculate corrections throughout the model.
To do so, we compute the one-dimensional Wasserstein distance between the target and the test-time distribution analytically by sorting the activations from both distributions. We subsequently utilize these distance measures for shifting the activations of the test-time distribution so that it matches the shape of the target distribution. Given that the target distribution is based on the training data, this effectively moves the test samples closer to data points seen during training and in doing so reduces the covariate shift. Consequently, the network can better process the features in subsequent layers. A subsequent step minimizes the difference to the original activation maps again and ensures that the proposed correction does not induce unwanted distortions.

In our experiments, we empirically show that our proposed method improves robustness against high-intensity noise of input corruptions. We evaluate and compare three normalization methods, i.e., BN, GN, and FRN on several standard image classification datasets, MNIST, CIFAR-10, ImageNet (ILSVRC 2012) and their corrupted variants.  Furthermore, we exemplary analyze the convergence behavior of the proposed correction and provide insights into the underlying principles. To summarize our contributions:
\begin{itemize}
    \item We propose an unsupervised non-parametric correction algorithm to mitigate distribution mismatches, caused by image corruptions, during test-time.
    \item In our experiments we analyze the impact of image corruptions in CNN architectures for different normalization methods using corrupted standard image classification datasets. 
    \item We provide insights into the convergence behavior and mechanisms responsible for the improved classification performance and empirically verify our assumptions. 
\end{itemize}

\section{Related work}\label{sec:related_work}
Deep learning methods achieve state-of-the-art performance for most machine learning benchmarks because of their flexibility and representational power. This flexibility, however, leads to over-fitting on the training data and thus reduces the robustness and generalization capabilities. Therefore, many methods -- such as weight regularization or dropout -- aim to reduce the problem of over-fitting~\cite{bishop2006pattern,JMLR:v15:srivastava14a}. Alternatively, robustness can be improved by increasing the coverage of the input space with data augmentation. Therefore training samples are augmented by the application of affine transformations or expected noise types and are then explicitly included in the training set~\cite{baird1992document,simard2003best,krizhevsky2012imagenet}. More elaborate methods optimize the augmentation via an additional neural network~\cite{cubuk2018autoaugment}. As mentioned before, improving robustness to one corruption type by data augmentation can lead to a decrease of robustness against others~\cite{pmlr-v97-gilmer19a,Benz_2021_WACV}.  
Other ways of improving robustness and prediction stability are representation learning techniques or capsule networks. These approaches try to learn equivariant representations of features; i.e., conceptual representations independent of the position, orientation or context~\cite{NEURIPS2020_d89a66c7,kim2017multi,zhou2017anomaly,ribeiro2020capsule}.
Moreover, the choice of activation functions impacts the robustness of the network as well~\cite{misra2019mish,nwankpa2018activation}.
%In recent years, much research focused on achieving adversarial robustness~\cite{goodfellow2014explaining}. Many methods aimed to provide defenses against these types of attacks, ranging from data prepossessing methods~\cite{buckman2018thermometer} over post-processing methods applied on the model output~\cite{197128}, to regularization methods that create smoother distributions within the network~\cite{miyato2015distributional}.
%Using smooth activation functions can also improve adversarial robustness~\cite{xie2020smooth}.
%As shown in~\cite{pmlr-v97-gilmer19a}, the effects of adversarial and Gaussian noise are inherently related; using a linear model approximation, they predict the distance of adversarial errors by the errors of Gaussian noise. This underscores that any measure aiming to improve adversarial robustness should further improve robustness against image corruptions.
Recently, there has also been increasing interest in improving the robustness of normalization methods. The influence of input corruptions on networks using batch normalization has been investigated in~\cite{Benz_2021_WACV}; this further led to a domain adaption method for the normalization that improves the robustness of DNNs against corruptions. This adaption can substantially improve the network's performance but requires a re-computation of the batch normalization parameters for each domain adaption. A similar approach was taken by~\cite{schneider2020improving}, which adapts the normalization parameters based on test-time statistics. Other approaches use self-supervised methods to retrain the model based on test samples to reduce domain shifts \cite{liang2020we,sun2020test,wang2021tent}.

\section{Distribution correction for Deep Neural Networks (DNNs)}\label{sec:noise_dnns}
Machine learning and signal processing tasks frequently experience performance degradation caused by noise. 
As noise comes in miscellaneous forms, it is challenging to achieve general robustness against arbitrary noise.
This is particularly problematic as the existence of noise introduces distribution mismatches (i.e., covariate shifts).
Thus, one promising direction for improving robustness is the reduction of such mismatches.
This is often tackled by means of normalizing the activations for each layer. Most existing approaches, however, struggle to do so as they are restricted to parametric distributions (e.g. Gaussians) that cannot correct for mismatches not reflected in the distribution parameters (i.e., the mean and variance).

In order to mitigate the distribution mismatches of the test-time activations $\mathbf{a}$, we must find an effective way to suppress noisy activations while maintaining the classification performance in the subsequent layers. Therefore, we will formulate this problem as a probabilistic denoising problem.
We first need to approximate the a-posteriori distribution,
\begin{equation}
\small
    p(\tilde{\mathbf{a}}|\mathbf{a}) = \dfrac{p(\mathbf{a}|\tilde{\mathbf{a}})p(\tilde{\mathbf{a}})}{p(\mathbf{a})},
\end{equation}
of the corrected activations $\tilde{\mathbf{a}}$ given the activations $\mathbf{a}$. 
Then, we can use the maximum a-posteriori estimate to determine the corrected activations; this is a well-established technique in image denoising~\cite{perona1990scale,buades2005review,rudin1992nonlinear,Zhu08anefficient}. 
For our considerations, we recast the maximum a-posteriori problem into an equivalent energy minimization problem to simplify the optimization procedure.  We assume that the prior, likelihood, and posterior come from an exponential family using a Gibbs measure so that 
\begin{equation}
\small
p(\tilde{\mathbf{a}}|\mathbf{a}) = e^{-\frac{E(\tilde{\mathbf{a}}|\mathbf{a})}{T}},\ 
    p(\tilde{\mathbf{a}}) = e^{-\frac{\mathcal{R}(\tilde{\mathbf{a}})}{T}},\ p(\mathbf{a}|\tilde{\mathbf{a}}) = e^{-\frac{\mathcal{D}(\mathbf{a}|\tilde{\mathbf{a}})}{T}}.\label{eq:exp_family}
\end{equation}
Note that we can omit the evidence term $p(\mathbf{a})$, as we do not perform model comparison. Then, by applying the logarithm to~\eqref{eq:exp_family}
 and by multiplying all terms with $-\frac{1}{T}$, we arrive at an energy minimization problem  
\begin{equation}
    \tilde{\mathbf{a}}^* = \argmin_{\tilde{\mathbf{a}}} E(\tilde{\mathbf{a}}|\mathbf{a}),
\end{equation}
with the optimal activation map $\tilde{\mathbf{a}}^*$ at its minimum.
The energy $E(\tilde{\mathbf{a}}|\mathbf{a})$ is composed of two terms $\mathcal{R}(\tilde{\mathbf{a}})$ (corresponding to the prior) and $D(\mathbf{a}|\tilde{\mathbf{a}})$ (corresponding to the likelihood) so that
\begin{equation}
\small
    E(\tilde{\mathbf{a}}|\mathbf{a}) = \mathcal{D}(\mathbf{a}|\tilde{\mathbf{a}})+\mathcal{R}(\tilde{\mathbf{a}}).
\end{equation}
For this form, we must, at the one hand, specify a suitable prior term $\mathcal{R}(\tilde{\mathbf{a}})$ that reduces the covariate shift in each layer without restricting the network (see Section \ref{sec:non_parametric_target}). The data likelihood term $\mathcal{D}(\mathbf{a}|\tilde{\mathbf{a}})$, on the other hand preserves the spatial correlations of the activation maps and prevents the loss of valuable information (see Section \ref{sec:likelihood}). By minimizing both terms jointly, one can achieve an optimal trade-off between minimizing the covariate shift and representing the available data.

\subsection{Including a non-parametric prior term}\label{sec:non_parametric_target}
Typically, parametric distributions do not provide good representations of the activation distributions in DNNs.
Consequently, any correction method that approximates the prior by a parametric target distribution $q_{\theta}(\mathbf{a})$ distorts the shape of the true activations distribution $p(\mathbf{a})$.
Subsequent layers are thus exposed to different input distributions than during training and suffer from the corresponding covariate shift. 
Ideally, the target distribution $q(\mathbf{a})$ should enforce similar (corrected) distributions as during training since any mismatch might outweigh the benefits of the noise-reduction otherwise.
If $q(\mathbf{a})$ should resemble the non-parametric distribution from the training set, however, it must be non-parametric as well.
The Wasserstein distance proves to be particularly well-suited for a novel correction method: not only does it allow to effectively minimize the mismatch between the distributions during training- and test-time, but it also provides an elegant way of representing a non-parametric distribution in one dimension.

To find a well-suited prior distribution, we must first sort the $N$ activations $\mathbf{a}$ for each sample $m$ in ascending order. Let $[\cdot]$ be the vector of all corresponding elements, then
\begin{equation}
\small
            [a^{(m)}_{(i)}],\mathbf{j} =\mathrm{sort}(\mathbf{a}^{(m)}),\\
\end{equation}\label{eq:sort}
where $a^{(m)}_{(i)}$ are the sorted activations with ${a}_{(i)}<{a}_{(i+1)}$, and $\mathbf{j}$ are the indices of the activations that are required for assigning activation updates. Note that the subscript $(i)$ always denotes sorted values. 

Second, to minimize the non-parametric prior term, we need to calculate the Wasserstein distance between the activation distribution during test-time (i.e., $p(\mathbf{a})$) and the prior. 
% (i.e., the Wasserstein barycenter $q(\mathbf{t})$ of the training samples). 
In order to create a useful target distribution, we require it to be stationary with respect to its general shape and location.
Unfortunately, this requirement prevents our method from considering channel-wise distributions, as the channel distributions are highly dependent on the input features. Therefore, we flatten the channels and create a single distribution across the height $H$, width $W$, and channel $C$ dimension of the layer resulting in $N=H\times W \times C$ activation values. In Appendix~\ref{sec:target_dist} we provide an empirical evaluation of the activation distribution with respect to stationarity.
As we do not care about the precise location of the distribution but primarily about its shape, we subtract the mean of the distributions so that
\begin{equation}
\small
     a'^{(m)}_{(i)} = a^{(m)}_{(i)} - \frac{1}{N} \sum_{i=1}^N a^{(m)}_{(i)}.
\end{equation}

Third, to construct the target values $ t_{(i)}$ (that represent the corrected activations $p(\tilde{\mathbf{a}})$) we utilize the Wasserstein barycenter, i.e., the distribution that minimizes the sum of the Wasserstein distances $W$ over all $M$ training distributions \cite{cuturi2014fast,anderes2016discrete}:
\begin{equation}
\small
    \min_q \sum_{m=1}^M W\big(q(\mathbf{a}),p(\mathbf{a})^{(m)}\big).
\end{equation}
In one dimension, the Wasserstein barycenter is simply the average over the order statistics of each sample $m$ so that 
\begin{equation}
\small
            t_{(i)} = \frac{1}{M} \sum_{m=1}^M a'^{(m)}_{(i)}.
\end{equation}
%We increase similarity by subtracting the mean of each distribution. Note that we do care about the precise location of the distribution but primarily about its shape. Consequently, we have
%\begin{equation}
%\small
%            t_{(i)} = t_{(i)} - \frac{1}{N} \sum_{i=1}^N t^{(m)}_{(i)}.
%\end{equation}
Finally, the one-dimensional Wasserstein distance between the target distribution $q(\mathbf{t})$ and the test-time distribution $p(\mathbf{a'}^{(m)})$ is given according to
\begin{equation}
\small
    W\big(p(\mathbf{a'}^{(m)}),q(\mathbf{t})\big) = \left(\sum_{i=1}^N||a'^{(m)}_{(i)}-t_{(i)}||^r\right)^{\frac{1}{r}}, \label{eq:distance}
\end{equation}
where $t_{(i)}$ are the sorted target values and $a'^{(m)}_{(i)}$ are the sorted test-time activations.\footnote{
As we aim for a scaleable correction method we will restrain from utilizing the labels $y$ in the form of a  conditional prior $p(\tilde{\mathbf{a}}|y)$ and restrict ourselves to using only a single distribution per layer.} 
Let $r=1$; then, the Wasserstein distance between $p(\mathbf{a'})$ and $q(\mathbf{t})$ from~\eqref{eq:distance} is minimized by updating the (unsorted) activation with index $j$ according to $\Delta_j = t_{(i)}-a'_{(i)}$.
%\begin{equation}
%    \Delta_j = t_{(i)}-a'_{(i)}.
%\end{equation}
We apply the correction after the ReLU activation function; thus many activations are zero. Our updates must preserves this sparsity as the performance will degrade otherwise (see the experiments in Section~\ref{sec:res_cifar10_corrupted}). Therefore, we explicitly enforce sparsity in the prior term $ \mathcal{R}$, i.e., we prevent correcting activations with $a=0$ by adding an infinitely deep energy-well $-\delta(a)$, where $\delta(\cdot)$ denotes the Dirac delta.\footnote{Note that this sparsity constraint is not required if the correction is applied directly after the convolution.}
Combining this sparsity term with the Wasserstein distance finally leads to the following prior term: 
\begin{equation}
\small
    \mathcal{R}(\mathbf{\tilde{a}}) = W\big(p(\mathbf{a'}^{(m)}),q(\mathbf{t})\big)- [\delta({a_i})].
\end{equation}\label{eq:prior_term}
This expression is straightforward to minimize according to
\begin{equation}
\small
\Delta_j =
\begin{cases}
 t_{(i)}-a'_{(i)} & \text{if $a\ne 0$,}\\
0 & \text{otherwise.} \label{eq:min_prior}
\end{cases}
\end{equation}

\subsection{Data likelihood}\label{sec:likelihood}
Minimizing only the prior term might have undesired side effects and destroy important structure in the channels of the network, i.e the spatial correlation. 
Therefore, the energy minimization needs to find a trade-off between matching the distributions and conserving the spatial correlation. We achieve this by considering a likelihood term $\mathcal{D}(\mathbf{a}|\tilde{\mathbf{a}})$, modeled by
\begin{equation}
\small
    \mathcal{D}(\mathbf{a}|\tilde{\mathbf{a}}) = \frac{1}{2}||\mathbf{a} -\tilde{\mathbf{a}}||^2
\end{equation}
that conserves the structure in the data. This expression is also straighforward to minimize by exploiting the gradient
\begin{equation}
\small
\dfrac{d\mathcal{D}}{d \mathbf{a}} = \mathbf{a} -\tilde{\mathbf{a}}. \label{eq:min_likelihood}
\end{equation}

\subsection{Correction algorithm}\label{sec:corr_algo}
Here we outline how to combine the prior and the likelihood term and how to minimize the energy $E(\tilde{\mathbf{a}}|\mathbf{a}) = \mathcal{D}(\mathbf{a}|\tilde{\mathbf{a}})+\mathcal{R}(\tilde{\mathbf{a}})$. Note that every prior-update modifies the corrected activations $\tilde{\mathbf{a}}$; thus we need to resort to an iterative procedure in which the likelihood and the prior term are alternately minimized according to~\eqref{eq:min_prior} and~\eqref{eq:min_likelihood}. 
Further implementation details are presented in the pseudocode in Appendix~\ref{sec:algo}.
Note that we choose independent values for the step-sizes of the prior-update, i.e., $\lambda_1$, and of the likelihood-update, i.e., $\lambda_2$, that implicitly determine the relative importance of the corresponding terms. In practice, the step-sizes should be chosen to achieve good classification performance (see the analysis in Section~\ref{sec:convergence}).

The proposed algorithm computes the corrections and successively minimizes the mismatch between the distributions $p(\mathbf{a}^{(m)})$ and $q(\mathbf{t})$ in a layer-wise fashion. That is -- starting from the input layer -- we correct the activations by performing the iterative updates. Further note, that we do not run the optimization procedure until convergence but stop the algorithm after $N_{iter}$ iterations instead; empirically we observe that only a few iterations are sufficient for improving the performance (see Section~\ref{sec:convergence}). 

As the correction is only applied at test-time, it can easily be retrofitted into existing models by adding the correction layer and calculating the target distributions $q(\mathbf{t})$ for each layer, using the training set.

%As convergence is hard to determine in a multi-layer CNN, will abort the algorithm after $n_{iter}$ iterations. As we show in Section \ref{sec:convergence}, only a small number of iterations are sufficient to achieve improved performance at a limited additional computational effort. Algorithm \ref{sec:algo} provides the implementation details of our proposed correction method.
%The update is then simply added to the corresponding activation $\tilde{a}_j = a_j + \Delta_j$, where $\tilde{a}$ is the updated activation value. This procedure minimizes the mismatch between distributions $p(a)$ with the target distribution $q(t)$.
%As the correction is only applied at test-time, it can be easily retrofitted into existing pre-trained networks by adding the correction layer and calculating the target values $t$ for the training set.
\section{Experiments}\label{sec:experiments}
In our experiments we will first exemplary analyze our proposed method with respect to its distribution matching capabilities (see Section \ref{sec:layers}), before we analyze the convergence behavior of the correction algorithm in Section \ref{sec:convergence}. In the remaining Sections \ref{sec:res_mnist_corrupted}, \ref{sec:res_cifar10_corrupted} and Section \ref{sec:res_imagenet} we present results for the corrupted classification datasets for MNIST, CIFAR-10 and ImageNet (ILSVRC 2012). All of the datasets are publicly available on TensorFlow datasets~\cite{mu2019mnist,hendrycks2018benchmarking}. We conduct the MNIST and CIFAR-10 experiments on a NVIDIA Tesla V100 and the ImageNet experiments on 4 NVIDIA RTX 2080Ti GPUs.
\subsection{Analyzing the effect of the correction within layers}\label{sec:layers}
\begin{figure}[h]
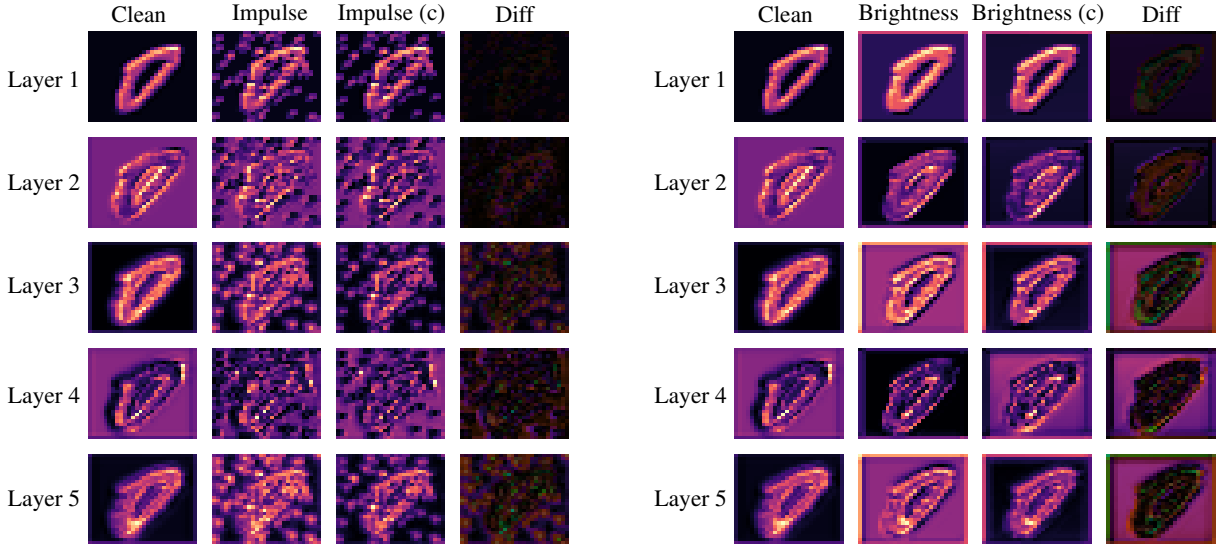

    \centering
    \begin{minipage}[h]{0.46\textwidth}
		\resizebox{\textwidth}{!}{
		\input{impulse.tikz}}
	\end{minipage}	
	\begin{minipage}[h]{0.05\textwidth}
	\hspace{\textwidth}
	\end{minipage}	
	\begin{minipage}[h]{0.46\textwidth}
		\resizebox{\textwidth}{!}{
		\input{brightness.tikz}}
	\end{minipage}	
    \caption{Comparison between the mean activation maps of the first 5 layers of a ResNet-20 trained on MNIST for two different corruption types. Impulse noise maps are shown on the left and Brightness corrupted maps on the right. (c) indicates the activation maps which use our proposed distribution correction and the Diff column shows the pixel-wise difference between corrected and uncorrected activation maps.}
    \label{fig:layer_images}
\end{figure}

Since the goal of this method is to reduce the covariate shift within the network, we analyze the activation maps and the distribution before and after the correction (denoted as (c)). The experimental details are presented in Section \ref{sec:res_mnist_corrupted}. In Figure \ref{fig:layer_images}, we see the difference between the corrected and uncorrected activation maps. When comparing the activation maps we always need to compare against the clean data variant. For the brightness corrupted inputs we see that the corrected layer is always close to the corresponding clean layer; this is especially visible in layers 3, 4, and 5. For the samples corrupted with impulse noise we see that for the network without correction, the noise seems to be amplified for the subsequent layers, whereas the corrected network is able to decrease the noise level. This indicates, that the network is operating in a region covered in training space. Applying the correction on the clean data, we observe that the modifications do not harm the activation map.
\begin{figure}[h]
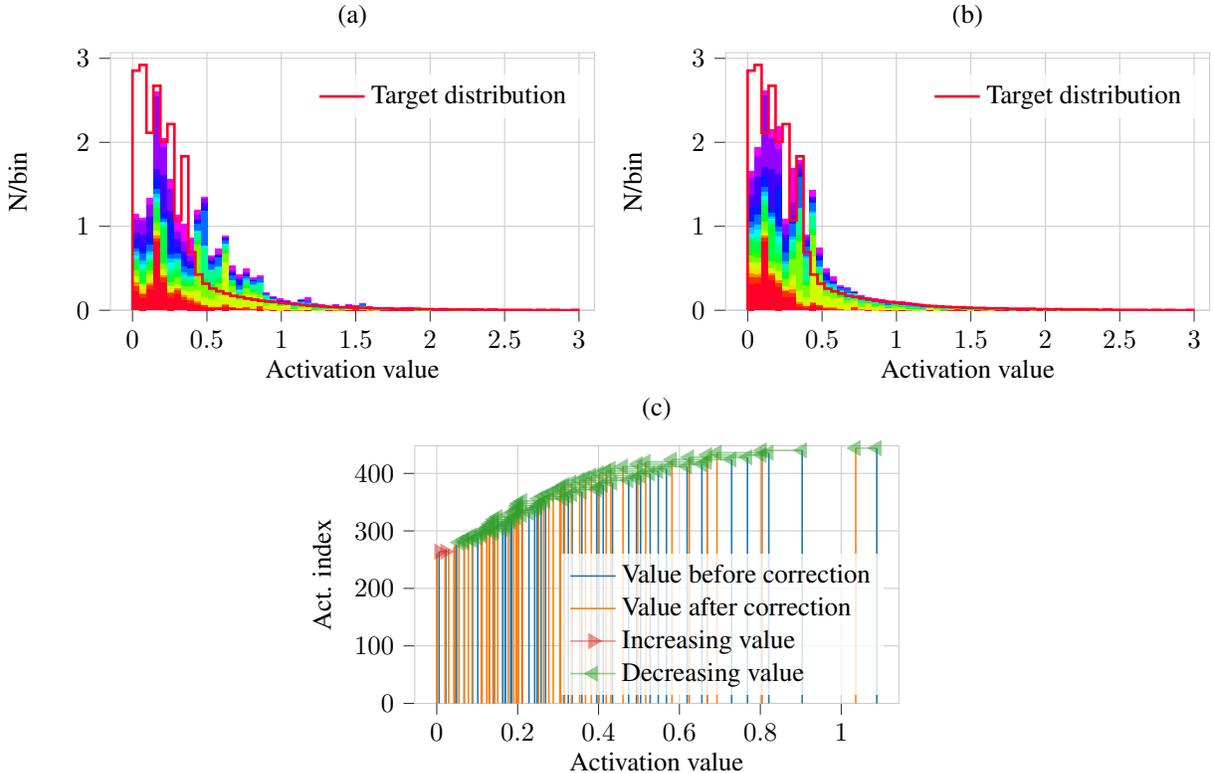

    \centering
	\begin{minipage}[h]{0.49\textwidth}
		\resizebox {\textwidth} {!} {
            \input{sample_1_before.tikz}}
    \end{minipage}
    	\begin{minipage}[h]{0.49\textwidth}
		\resizebox {\textwidth} {!} {
            \input{sample_1_after.tikz}}
    \end{minipage}
    \begin{minipage}[h]{0.49\textwidth}
		\resizebox {\textwidth} {!} {
            \input{sample_1_lines.tikz}}
    \end{minipage}
    \caption{(a) and (b) show the stacked channel histograms (excluding zeros) of the activations of the $1^{st}$ layer of a ResNet-20 for a MNIST sample containing impulse noise, before (a) and after (b) the Wasserstein correction. Different colors indicate activation histograms from different channels. In (c) the activation values before and after correction are depicted.}    \label{fig:distributions_before_after}
\end{figure}
In Figure~\ref{fig:distributions_before_after}, we show the impact of the correction on the distribution of activations for a noisy MNIST sample. In the before-image Figure~\ref{fig:distributions_before_after}~(a), the activations are further away from the target distribution (red silhouette) than after the correction Figure~\ref{fig:distributions_before_after}~(b). This shows that, while we do not exactly match the distribution (as $\lambda_2 > 0$), we are able to effectively reduce the covariate shift. In Figure~\ref{fig:distributions_before_after}~(c), we see that most updates are in the same direction, with only values close to zero being pushed towards larger values. If we assume that high-impact noise comes from the tails of the distribution this also reduces the overall noise level. 
\subsection{Convergence behavior}\label{sec:convergence}
As our correction method is applied iteratively, we investigate its convergence behavior for the average classification accuracy on the corrupted MNIST dataset. Here we vary the step size parameters $\lambda_1$ and $\lambda_2$ and show their influence.
\begin{figure}[h]
    \centering
	\begin{minipage}[h]{0.49\textwidth}
		\resizebox{\textwidth}{!}{
		% This file was created by tikzplotlib v0.9.6.
\begin{tikzpicture}

\definecolor{color0}{rgb}{0.12156862745098,0.466666666666667,0.705882352941177}
\definecolor{color1}{rgb}{1,0.498039215686275,0.0549019607843137}
\definecolor{color2}{rgb}{0.172549019607843,0.627450980392157,0.172549019607843}
\definecolor{color3}{rgb}{0.83921568627451,0.152941176470588,0.156862745098039}

\begin{axis}[height=5cm,width=8cm,
axis line style={white!80!black},
legend cell align={left},
legend style={fill opacity=0.8,font=\small, draw opacity=1, text opacity=1, at={(0.97,0.03)}, anchor=south east, draw=none},
tick align=outside,
tick pos=left,
title={Batch norm},
x grid style={white!80!black},
xmajorgrids,
xmin=-0.25, xmax=5.25,
xtick style={color=white!15!black},
y grid style={white!80!black},
ymajorgrids,
xlabel={Iteration},
ylabel={Accuracy},
ymin=0.742396566737443, ymax=0.857972186338156,
ytick style={color=white!15!black}
]

%\addlegendentry{bn_lamb_0.25_0.5}
\addplot [semithick, color0, mark=*, mark size=2, mark options={solid}]
table {%
0 0.747650003992021
1 0.786168745718896
2 0.852718749083579
3 0.848174996674061
4 0.847024995833635
5 0.846656247973442
};
%\addlegendentry{bn_lamb_0.25_0.5}
\addplot [semithick, color1, mark=*, mark size=2, mark options={solid}]
table {%
0 0.747650003992021
1 0.765349995344877
2 0.788268745876849
3 0.795199999585748
4 0.796912492252886
5 0.797374993562698
};
%\addlegendentry{bn_lamb_0.25_0.5}
\addplot [semithick, color2, mark=*, mark size=2, mark options={solid}]
table {%
0 0.747650003992021
1 0.772537497803569
2 0.831656253896654
3 0.841462500393391
4 0.842181256040931
5 0.84205000475049
};
%\addlegendentry{bn_lamb_0.25_0.5}
\addplot [semithick, color3, mark=*, mark size=2, mark options={solid}]
table {%
0 0.747650003992021
1 0.759056253358722
2 0.766175001859665
3 0.769824991934001
4 0.771056246943772
5 0.771500001661479
};

\addlegendentry{$\lambda_1=0.75$, $\lambda_2=0.25$}
\addlegendentry{$\lambda_1=0.5$, $\lambda_2=0.5$}
\addlegendentry{$\lambda_1=0.5$, $\lambda_2=0.25$}
\addlegendentry{$\lambda_1=0.25$, $\lambda_2=0.5$}

\addplot [semithick, color0, mark=+, mark size=5, mark options={solid}, only marks]
table {%
0 0.747650003992021
1 0.786168745718896
2 0.852718749083579
3 0.848174996674061
4 0.847024995833635
5 0.846656247973442
};
%\addlegendentry{bn_lamb_0.25_0.5}
\addplot [semithick, color0, mark=+, mark size=5, mark options={solid}, only marks]
table {%
0 0.747650003992021
1 0.786168745718896
2 0.852718749083579
3 0.848174996674061
4 0.847024995833635
5 0.846656247973442
};
%\addlegendentry{bn_lamb_0.25_0.5}
\addplot [semithick, color1, mark=+, mark size=5, mark options={solid}, only marks]
table {%
0 0.747650003992021
1 0.765349995344877
2 0.788268745876849
3 0.795199999585748
4 0.796912492252886
5 0.797374993562698
};
%\addlegendentry{bn_lamb_0.25_0.5}
\addplot [semithick, color1, mark=+, mark size=5, mark options={solid}, only marks]
table {%
0 0.747650003992021
1 0.765349995344877
2 0.788268745876849
3 0.795199999585748
4 0.796912492252886
5 0.797374993562698
};
%\addlegendentry{bn_lamb_0.25_0.5}
\addplot [semithick, color2, mark=+, mark size=5, mark options={solid}, only marks]
table {%
0 0.747650003992021
1 0.772537497803569
2 0.831656253896654
3 0.841462500393391
4 0.842181256040931
5 0.84205000475049
};
%\addlegendentry{bn_lamb_0.25_0.5}
\addplot [semithick, color2, mark=+, mark size=5, mark options={solid}, only marks]
table {%
0 0.747650003992021
1 0.772537497803569
2 0.831656253896654
3 0.841462500393391
4 0.842181256040931
5 0.84205000475049
};
%\addlegendentry{bn_lamb_0.25_0.5}
\addplot [semithick, color3, mark=+, mark size=5, mark options={solid}, only marks]
table {%
0 0.747650003992021
1 0.759056253358722
2 0.766175001859665
3 0.769824991934001
4 0.771056246943772
5 0.771500001661479
};
%\addlegendentry{bn_lamb_0.25_0.5}
\addplot [semithick, color3, mark=+, mark size=5, mark options={solid}, only marks]
table {%
0 0.747650003992021
1 0.759056253358722
2 0.766175001859665
3 0.769824991934001
4 0.771056246943772
5 0.771500001661479
};

%\addlegendentry{bn_lamb_0.25_0.5}
\end{axis}

\end{tikzpicture}}
    \end{minipage}
    	\begin{minipage}[h]{0.49\textwidth}
		\resizebox{\textwidth}{!}{
		% This file was created by tikzplotlib v0.9.6.
\begin{tikzpicture}

\definecolor{color1}{rgb}{0.12156862745098,0.466666666666667,0.705882352941177}%blue
\definecolor{color0}{rgb}{1,0.498039215686275,0.0549019607843137}%orange
\definecolor{color3}{rgb}{0.172549019607843,0.627450980392157,0.172549019607843}%green
\definecolor{color2}{rgb}{0.83921568627451,0.152941176470588,0.156862745098039}%red

\begin{axis}[height=5cm,width=8cm,
axis line style={white!80!black},
legend cell align={left},
legend style={fill opacity=0.8,font=\small, draw opacity=1, text opacity=1, at={(0.97,0.03)}, anchor=south east, draw=none},
tick align=outside,
tick pos=left,
title={Filter response norm},
x grid style={white!80!black},
xmajorgrids,
xlabel={Iteration},
ylabel={Accuracy},
xmin=-0.25, xmax=5.25,
xtick style={color=white!15!black},
y grid style={white!80!black},
ymajorgrids,
ymin=0.909759686328471, ymax=0.938559058122337,
ytick style={color=white!15!black}
]

\addplot [semithick, color1, mark=*, mark size=2, mark options={solid}]
table {%
0 0.911068748682737
1 0.936662502586842
2 0.930531248450279
3 0.927750006318092
4 0.927131246775389
5 0.927062503993511
};
%\addlegendentry{frn_lamb_0.5_0.25}
\addplot [semithick, color0, mark=*, mark size=2, mark options={solid}]
table {%
0 0.911068748682737
1 0.937206242233515
2 0.935612495988607
3 0.934862498193979
4 0.934618748724461
5 0.934568747878075
};
%\addlegendentry{frn_lamb_0.5_0.25}
%\addlegendentry{frn_lamb_0.5_0.25}
\addplot [semithick, color3, mark=*, mark size=2, mark options={solid}]
table {%
0 0.911068748682737
1 0.937125004827976
2 0.93368124961853
3 0.931637495756149
4 0.930443748831749
5 0.929968751966953
};
%\addlegendentry{frn_lamb_0.5_0.25}
\addplot [semithick, color2, mark=*, mark size=2, mark options={solid}]
table {%
0 0.911068748682737
1 0.93724999576807
2 0.936893753707409
3 0.936456248164177
4 0.936387501657009
5 0.936431251466274
};

\addlegendentry{$\lambda_1=0.75$, $\lambda_2=0.25$}
\addlegendentry{$\lambda_1=0.5$, $\lambda_2=0.5$}
\addlegendentry{$\lambda_1=0.5$, $\lambda_2=0.25$}
\addlegendentry{$\lambda_1=0.25$, $\lambda_2=0.5$}

\addplot [semithick, color0, mark=+, mark size=5, mark options={solid}, only marks]
table {%
0 0.911068748682737
1 0.937206242233515
2 0.935612495988607
3 0.934862498193979
4 0.934618748724461
5 0.934568747878075
};
%\addlegendentry{frn_lamb_0.5_0.25}
\addplot [semithick, color0, mark=+, mark size=5, mark options={solid}, only marks]
table {%
0 0.911068748682737
1 0.937206242233515
2 0.935612495988607
3 0.934862498193979
4 0.934618748724461
5 0.934568747878075
};
%\addlegendentry{frn_lamb_0.5_0.25}
\addplot [semithick, color1, mark=+, mark size=5, mark options={solid}, only marks]
table {%
0 0.911068748682737
1 0.936662502586842
2 0.930531248450279
3 0.927750006318092
4 0.927131246775389
5 0.927062503993511
};
%\addlegendentry{frn_lamb_0.5_0.25}
\addplot [semithick, color1, mark=+, mark size=5, mark options={solid}, only marks]
table {%
0 0.911068748682737
1 0.936662502586842
2 0.930531248450279
3 0.927750006318092
4 0.927131246775389
5 0.927062503993511
};
%\addlegendentry{frn_lamb_0.5_0.25}
\addplot [semithick, color2, mark=+, mark size=5, mark options={solid}, only marks]
table {%
0 0.911068748682737
1 0.93724999576807
2 0.936893753707409
3 0.936456248164177
4 0.936387501657009
5 0.936431251466274
};
%\addlegendentry{frn_lamb_0.5_0.25}
\addplot [semithick, color2, mark=+, mark size=5, mark options={solid}, only marks]
table {%
0 0.911068748682737
1 0.93724999576807
2 0.936893753707409
3 0.936456248164177
4 0.936387501657009
5 0.936431251466274
};
%\addlegendentry{frn_lamb_0.5_0.25}
\addplot [semithick, color3, mark=+, mark size=5, mark options={solid}, only marks]
table {%
0 0.911068748682737
1 0.937125004827976
2 0.93368124961853
3 0.931637495756149
4 0.930443748831749
5 0.929968751966953
};
%\addlegendentry{frn_lamb_0.5_0.25}
\addplot [semithick, color3, mark=+, mark size=5, mark options={solid}, only marks]
table {%
0 0.911068748682737
1 0.937125004827976
2 0.93368124961853
3 0.931637495756149
4 0.930443748831749
5 0.929968751966953
};

%\addlegendentry{frn_lamb_0.5_0.25}
\end{axis}

\end{tikzpicture}}
    \end{minipage}
    \begin{minipage}[h]{0.49\textwidth}
		\resizebox{\textwidth}{!}{
		% This file was created by tikzplotlib v0.9.6.
\begin{tikzpicture}

\definecolor{color3}{rgb}{0.12156862745098,0.466666666666667,0.705882352941177}%blue
\definecolor{color0}{rgb}{1,0.498039215686275,0.0549019607843137}%orange
\definecolor{color2}{rgb}{0.172549019607843,0.627450980392157,0.172549019607843}%green
\definecolor{color1}{rgb}{0.83921568627451,0.152941176470588,0.156862745098039}%red

\begin{axis}[height=5cm,width=8cm,
axis line style={white!80!black},
legend cell align={left},
legend style={fill opacity=0.8,font=\small, draw opacity=1, text opacity=1, at={(0.97,0.03)}, anchor=south east, draw=none},
tick align=outside,
tick pos=left,
x grid style={white!80!black},
xmajorgrids,
title={Group norm},
xlabel={Iteration},
ylabel={Accuracy},
xmin=-0.25, xmax=5.25,
xtick style={color=white!15!black},
y grid style={white!80!black},
ymajorgrids,
ymin=0.926632500626147, ymax=0.936642502062023,
ytick style={color=white!15!black}
]

\addplot [semithick, color3, mark=*, mark size=2, mark options={solid}]
table {%
0 0.933893751353025
1 0.934668753296137
2 0.929424993693829
3 0.927512504160404
4 0.92713749781251
5 0.927087500691414
};
%\addlegendentry{gn_lamb_0.75_0.25}
\addplot [semithick, color0, mark=*, mark size=2, mark options={solid}]
table {%
0 0.933893751353025
1 0.936187501996756
2 0.935212504118681
3 0.934850007295609
4 0.934818755835295
5 0.934806257486343
};
\addplot [semithick, color2, mark=*, mark size=2, mark options={solid}]
table {%
0 0.933893751353025
1 0.935906246304512
2 0.933131247758865
3 0.931568760424852
4 0.930643763393164
5 0.9302750043571
};
%\addlegendentry{gn_lamb_0.75_0.25}
\addplot [semithick, color1, mark=*, mark size=2, mark options={solid}]
table {%
0 0.933893751353025
1 0.935456249862909
2 0.935781251639128
3 0.935687500983477
4 0.935612499713898
5 0.935562498867512
};
%\addlegendentry{gn_lamb_0.75_0.25}

%\addlegendentry{gn_lamb_0.75_0.25}

\addlegendentry{$\lambda_1=0.75$, $\lambda_2=0.25$}
\addlegendentry{$\lambda_1=0.5$, $\lambda_2=0.5$}
\addlegendentry{$\lambda_1=0.5$, $\lambda_2=0.25$}
\addlegendentry{$\lambda_1=0.25$, $\lambda_2=0.5$}

\addplot [semithick, color0, mark=+, mark size=5, mark options={solid}, only marks]
table {%
0 0.933893751353025
1 0.936187501996756
2 0.935212504118681
3 0.934850007295609
4 0.934818755835295
5 0.934806257486343
};
%\addlegendentry{gn_lamb_0.75_0.25}
\addplot [semithick, color0, mark=+, mark size=5, mark options={solid}, only marks]
table {%
0 0.933893751353025
1 0.936187501996756
2 0.935212504118681
3 0.934850007295609
4 0.934818755835295
5 0.934806257486343
};
%\addlegendentry{gn_lamb_0.75_0.25}
\addplot [semithick, color1, mark=+, mark size=5, mark options={solid}, only marks]
table {%
0 0.933893751353025
1 0.935456249862909
2 0.935781251639128
3 0.935687500983477
4 0.935612499713898
5 0.935562498867512
};
%\addlegendentry{gn_lamb_0.75_0.25}
\addplot [semithick, color1, mark=+, mark size=5, mark options={solid}, only marks]
table {%
0 0.933893751353025
1 0.935456249862909
2 0.935781251639128
3 0.935687500983477
4 0.935612499713898
5 0.935562498867512
};
%\addlegendentry{gn_lamb_0.75_0.25}
\addplot [semithick, color2, mark=+, mark size=5, mark options={solid}, only marks]
table {%
0 0.933893751353025
1 0.935906246304512
2 0.933131247758865
3 0.931568760424852
4 0.930643763393164
5 0.9302750043571
};
%\addlegendentry{gn_lamb_0.75_0.25}
\addplot [semithick, color2, mark=+, mark size=5, mark options={solid}, only marks]
table {%
0 0.933893751353025
1 0.935906246304512
2 0.933131247758865
3 0.931568760424852
4 0.930643763393164
5 0.9302750043571
};
%\addlegendentry{gn_lamb_0.75_0.25}
\addplot [semithick, color3, mark=+, mark size=5, mark options={solid}, only marks]
table {%
0 0.933893751353025
1 0.934668753296137
2 0.929424993693829
3 0.927512504160404
4 0.92713749781251
5 0.927087500691414
};
%\addlegendentry{gn_lamb_0.75_0.25}
\addplot [semithick, color3, mark=+, mark size=5, mark options={solid}, only marks]
table {%
0 0.933893751353025
1 0.934668753296137
2 0.929424993693829
3 0.927512504160404
4 0.92713749781251
5 0.927087500691414
};

\end{axis}

\end{tikzpicture}}
    \end{minipage}
    \caption{Performance over the number of used iterations for the correction algorithm for a single model trained on MNIST. Each normalization method uses 4 sets of step size parameters $\lambda_1$ and $\lambda_2$.}\label{fig:iterations}
\end{figure}
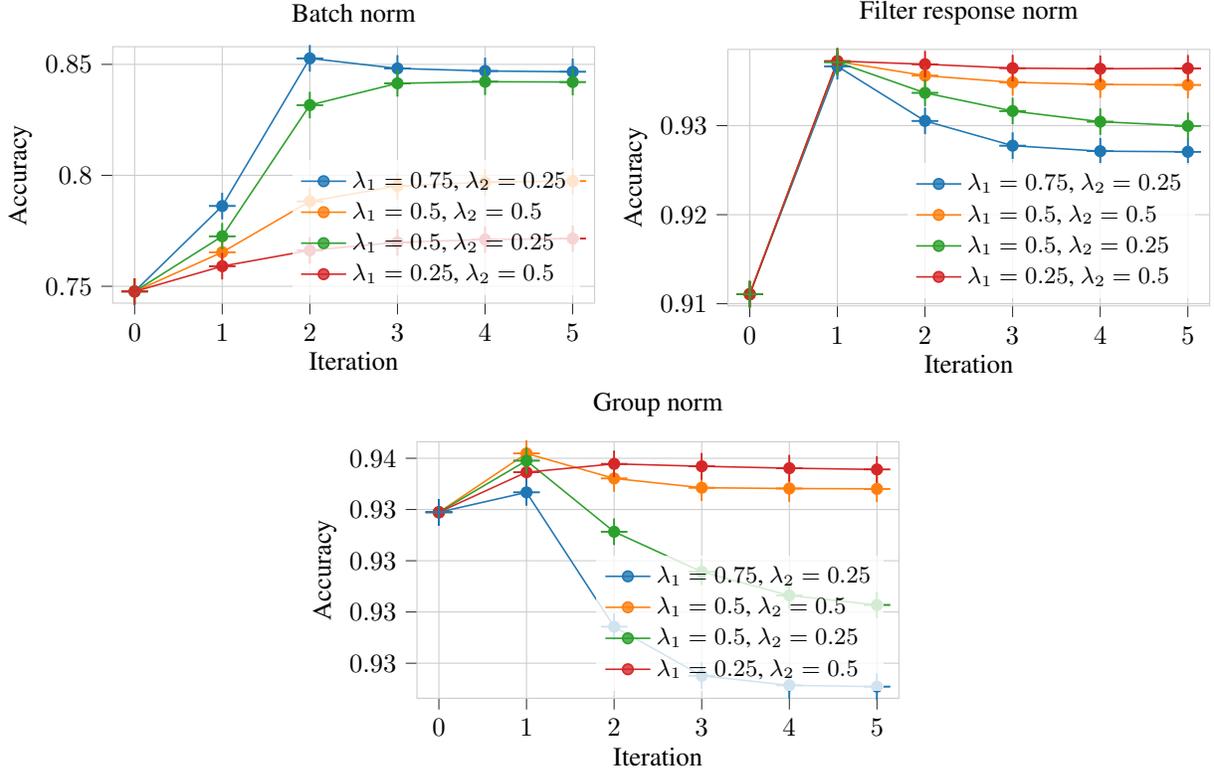
In Figure \ref{fig:iterations} we see that BN, FRN, and GN, show significantly different behavior for the same choices of $\lambda_1$ and $\lambda_2$. FRN and GN obtain substantially better results than BN even without using the correction (at iteration 0). Also, BN is the only model that can substantially improve its classification performance by running the algorithm for more than 1 iteration. FRN shows consistent results after one iteration, whereas the performances diverge for more iterations. GN also shows improvement within the second iteration for the parameter set $\lambda_1 = 0.25,\ \lambda_2=0.5$, but does not outperform the best model obtained with only one iteration.

\subsection{Corrupted MNIST classification}\label{sec:res_mnist_corrupted}
The corrupted MNIST dataset contains 15 different corruption variants of the original MNIST images (see \cite{mu2019mnist} for details). It contains 10000 gray scale images of size 28$\times$28 per corruption, which we used for our evaluations. 
For this experiment we choose different sets of steps size parameters $\lambda_1$ and $\lambda_2$ for each normalization method, which are listed in the Appendix in Table \ref{tab:mnist_set}. First, we trained 10 randomly initialized ResNet-20 models for 50 epochs using an SGD optimizer with a base learning rate of 0.1 on the clean MNIST dataset \cite{LeCun2010}. We decayed the learning rate after 25 and 40 epochs. The input data was normalized to a range $[0,1]$. The trained networks were then evaluated using the corrupted data, with and without the correction (c). For the corrected variant, the target distribution of the MNIST training set is required.
\begin{table}[]
\tiny
    \centering
        \caption{Classification accuracy in $\%$ on the corrupted MNIST dataset without and with the proposed distribution correction method (c) for BN, FRN and GN.}
    \begin{tabular}{c||c|c||c|c||c|c}
         noise type&BN&BN (c)&FRN&FRN (c)&GN&GN (c)\\
         \hline\hline 
 identity&99.54$\pm$0.04&99.21$\pm$0.2&99.51$\pm$0.06&99.51$\pm$0.05&99.6$\pm$0.07&\textbf{99.6$\pm$0.06} \\ \hline
 shot noise&97.66$\pm$0.26&96.42$\pm$0.8&98.28$\pm$0.24&\textbf{98.3$\pm$0.24}&98.17$\pm$0.22&98.16$\pm$0.22 \\ \hline
 impulse noise&37.86$\pm$5.3&55.4$\pm$11.36&92.27$\pm$1.83&\textbf{93.08$\pm$1.24}&92.3$\pm$1.46&92.68$\pm$1.4 \\ \hline
 glass blur&77.34$\pm$5.41&72.26$\pm$10.07&92.94$\pm$0.93&93.27$\pm$0.91&\textbf{93.71$\pm$0.6}&93.57$\pm$0.63 \\ \hline
 motion blur&96.26$\pm$1.07&95.93$\pm$1.41&98.28$\pm$0.27&98.23$\pm$0.26&\textbf{98.79$\pm$0.07}&98.75$\pm$0.09 \\ \hline
 shear&98.98$\pm$0.12&98.03$\pm$0.73&99.04$\pm$0.06&99.03$\pm$0.07&\textbf{99.24$\pm$0.07}&99.23$\pm$0.07 \\ \hline
 scale&97.63$\pm$0.26&97.06$\pm$0.84&98.02$\pm$0.2&97.99$\pm$0.26&98.13$\pm$0.2&\textbf{98.22$\pm$0.21} \\ \hline
 rotate&\textbf{95.67$\pm$0.33}&94.66$\pm$0.87&95.33$\pm$0.43&95.38$\pm$0.4&95.45$\pm$0.42&95.46$\pm$0.37 \\ \hline
 brightness&26.09$\pm$9.18&62.59$\pm$27.52&94.15$\pm$3.58&99.1$\pm$0.24&99.41$\pm$0.08&\textbf{99.41$\pm$0.08} \\ \hline
 translate&98.45$\pm$0.2&97.68$\pm$0.84&94.74$\pm$1.89&94.45$\pm$1.98&98.99$\pm$0.13&\textbf{99.0$\pm$0.14} \\ \hline
 stripe&21.37$\pm$4.82&24.64$\pm$8.2&\textbf{79.91$\pm$7.9}&73.07$\pm$10.84&44.95$\pm$14.5&46.95$\pm$14.75 \\ \hline
 fog&20.42$\pm$6.09&54.68$\pm$25.14&77.0$\pm$8.61&96.76$\pm$1.39&97.26$\pm$1.46&\textbf{97.42$\pm$1.2} \\ \hline
 spatter&96.73$\pm$0.56&95.59$\pm$1.06&\textbf{97.64$\pm$0.2}&97.51$\pm$0.19&97.53$\pm$0.25&97.48$\pm$0.27 \\ \hline
 dotted line&96.18$\pm$1.12&93.65$\pm$2.82&\textbf{97.26$\pm$0.81}&97.04$\pm$0.79&96.59$\pm$0.65&96.43$\pm$0.7 \\ \hline
 zigzag&77.02$\pm$1.79&73.42$\pm$6.13&88.31$\pm$0.69&\textbf{88.41$\pm$0.62}&87.12$\pm$0.98&87.12$\pm$0.96 \\ \hline
 canny edges&69.4$\pm$7.39&73.19$\pm$6.64&\textbf{83.14$\pm$3.65}&83.1$\pm$3.85&78.26$\pm$1.4&77.58$\pm$1.33 \\ \hline \hline
 \textbf{average}&76.83$\pm$0.8&81.39$\pm$5.02&93.25$\pm$1.21&\textbf{94.34$\pm$0.8}&92.65$\pm$0.9&92.74$\pm$0.94 \\ 
    \end{tabular}
    \label{tab:mnist}
\end{table}
The results in Table \ref{tab:mnist} show the average accuracies of the 10 randomly initialized models for the different corruption types. We see that our proposed correction method generally improves classification accuracy. Especially for BN, we see that our method can substantially improve the average classification performance. This is not unexpected, as BN is the most vulnerable normalization method with respect to distribution changes. Here we also see that there is a large variance of $\sigma=5.02$ between the corrected results of the different models, indicating that not all networks converged with the chosen step size parameter set $\lambda_1$ and $\lambda_2$. The best overall results were achieved using the corrected FRN models achieving $94.34\pm 0.8~\%$ accuracy over all models and corruption variants, achieving a performance improvement of $17.51~\%$, compared to the standard BN models. GN performed similarly with and without the correction method. 
\subsection{Corrupted CIFAR-10 classification}\label{sec:res_cifar10_corrupted}
The corrupted CIFAR-10 dataset contains 19 different corruption variants of the original dataset (see \cite{hendrycks2018benchmarking} for details). The corrupted CIFAR-10 dataset additionally features 5 different levels of corruption severity for each corruption type. It contains 10000 RGB images of size 32$\times$32 per corruption type and severity. 
For the evaluation on the corrupted CIFAR-10 classification task we choose the same parameter set $\lambda_1=1.0$ and $\lambda_2=0.2$ and $N_{iter}=1$ for all normalization methods. We trained 10 randomly initialized ResNet-20 models for 300 epochs using an SGD optimizer with a base learning rate of 0.1 on the clean CIFAR-10 dataset \cite{CIFAR-10}. During training, we decayed the learning rate after 150 and 225 epochs by a factor of 0.1. The input data was normalized to zero mean and unit variance and a widely used standard data augmentation scheme was performed~\cite{he2016deep,DBLP:journals/corr/HuangLW16a}.
%As for MNIST the trained networks, we then evaluated the models using the corrupted dataset, which required the acquisition of the target distributions for the corrected networks using the clean training set. 
\begin{table}[h]
\tiny
    \centering
        \caption{Average classification accuracy in $\%$ on the corrupted CIFAR-10 dataset without and with the proposed distribution correction method (c) for BN, FRN and GN.}
    \begin{tabular}{c||c|c||c|c||c|c}
         noise type&BN&BN (c)&FRN&FRN (c)&GN&GN (c)\\
         \hline\hline 
 brightness&\textbf{90.05}&89.19&86.6&86.09&87.95&87.29\\ \hline
 contrast&73.85&82.95&83.55&83.41&\textbf{88.18}&87.71\\ \hline
 defocus blur&79.06&84.69&81.69&81.44&\textbf{84.85}&84.27\\ \hline
 elastic&79.90&80.1&78.01&77.48&\textbf{81.25}&80.31\\ \hline
 fog&84.14&84.75&83.0&82.54&\textbf{84.85}&84.17\\ \hline
 frost&73.82&76.8&75.73&75.58&79.13&\textbf{79.23} \\ \hline
 frosted glass blur&53.82&56.01&56.45&56.54&\textbf{60.47}&60.38\\ \hline
 gaussian blur&69.82&78.98&76.35&76.28&\textbf{81.2}&80.86\\ \hline
 gaussian noise&51.2&58.87&60.92&62.38&60.58&\textbf{63.28} \\ \hline
 impulse noise&60.72&66.29&67.14&67.49&68.67&\textbf{68.77} \\ \hline
 jpeg compression&\textbf{77.79}&74.41&73.13&72.45&76.73&75.49\\ \hline
 motion blur&72.8&78.32&78.51&78.23&\textbf{82.98}&82.41\\ \hline
 pixelate&70.87&71.55&71.57&71.08&\textbf{75.4}&74.8\\ \hline
 saturate&\textbf{88.64}&87.45&85.39&84.8&86.82&86.11\\ \hline
 shot noise&63.51&69.54&70.15&70.93&69.42&\textbf{71.09} \\ \hline
 snow&76.70&77.37&75.86&75.41&\textbf{78.95}&78.64\\ \hline
 spatter&80.57&\textbf{82.19}&80.05&79.84&81.33&81.21\\ \hline
 speckle noise&66.06&70.58&70.97&\textbf{71.48}&69.89&71.06\\ \hline
 zoom blur&72.62&80.23&75.83&75.62&\textbf{80.97}&80.26\\ \hline
 identity&\textbf{91.67}&90.46&87.8&87.29&89.05&88.37\\ \hline \hline
 \textbf{avg. accuracy} &72.95&76.33&75.31&75.21&\textbf{77.88}&77.75\\ 
    \end{tabular}
    \label{tab:cifar10}
\end{table}
\begin{figure}[h]
\begin{minipage}[h]{0.49\textwidth}
    \centering
		\resizebox{\textwidth}{!}{
		% This file was created by tikzplotlib v0.9.6.
\begin{tikzpicture}

\definecolor{color0}{rgb}{0.12156862745098,0.466666666666667,0.705882352941177}
\definecolor{color3}{rgb}{0.172549019607843,0.627450980392157,0.172549019607843}
\definecolor{color4}{rgb}{0.172549019607843,0.627450980392157,0.172549019607843}
\definecolor{color5}{rgb}{0.83921568627451,0.152941176470588,0.156862745098039}
\definecolor{color1}{rgb}{0.12156862745098,0.466666666666667,0.705882352941177}
\definecolor{color2}{rgb}{0.83921568627451,0.152941176470588,0.156862745098039}

\begin{axis}[height=5cm,width=8cm,
axis line style={white!80!black},
legend cell align={left},
legend style={fill opacity=0.8,font=\small, draw opacity=1, text opacity=1, at={(0.4,0.03)}, anchor=south east, draw=none},
tick align=outside,
tick pos=left,
x grid style={white!80!black},
xmajorgrids,
xlabel={Severity},
ylabel={Accuracy},
xmin=-0.25, xmax=5.25,
xtick style={color=white!15!black},
y grid style={white!80!black},
ymajorgrids,
ymin=0.483475111622699, ymax=0.94015639056,
ytick style={color=white!15!black}
]
\path [draw=color0, semithick]
(axis cs:0,0.914001860454317)
--(axis cs:0,0.919398150608305);

\path [draw=color0, semithick]
(axis cs:1,0.831338594492065)
--(axis cs:1,0.839560817247597);

\path [draw=color0, semithick]
(axis cs:2,0.76454012639293)
--(axis cs:2,0.775868058958034);

\path [draw=color0, semithick]
(axis cs:3,0.699074955046524)
--(axis cs:3,0.715279428898941);

\path [draw=color0, semithick]
(axis cs:4,0.622823893476053)
--(axis cs:4,0.642220547175445);

\path [draw=color0, semithick]
(axis cs:5,0.504233351574395)
--(axis cs:5,0.52506138550111);

\path [draw=color1, semithick]
(axis cs:0,0.901898823017998)
--(axis cs:0,0.907323398356513);

\path [draw=color1, semithick]
(axis cs:1,0.834916205863073)
--(axis cs:1,0.840818298758555);

\path [draw=color1, semithick]
(axis cs:2,0.788264029803453)
--(axis cs:2,0.795303221784019);

\path [draw=color1, semithick]
(axis cs:3,0.74386839747276)
--(axis cs:3,0.754066105692145);

\path [draw=color1, semithick]
(axis cs:4,0.68539682037221)
--(axis cs:4,0.701161075514061);

\path [draw=color1, semithick]
(axis cs:5,0.592762051937106)
--(axis cs:5,0.613862509471701);

\path [draw=color2, semithick]
(axis cs:0,0.875333635946893)
--(axis cs:0,0.892046363213874);

\path [draw=color2, semithick]
(axis cs:1,0.818710570898417)
--(axis cs:1,0.840006270849469);

\path [draw=color2, semithick]
(axis cs:2,0.783357105936799)
--(axis cs:2,0.806953416820279);

\path [draw=color2, semithick]
(axis cs:3,0.754611732146325)
--(axis cs:3,0.778627218382271);

\path [draw=color2, semithick]
(axis cs:4,0.715023054683454)
--(axis cs:4,0.738964312532456);

\path [draw=color2, semithick]
(axis cs:5,0.651551841603456)
--(axis cs:5,0.675359736449466);

\path [draw=color3, semithick]
(axis cs:0,0.868294253135301)
--(axis cs:0,0.877485738014601);

\path [draw=color3, semithick]
(axis cs:1,0.805916305738573)
--(axis cs:1,0.817683694090769);

\path [draw=color3, semithick]
(axis cs:2,0.763435243269797)
--(axis cs:2,0.780474229271008);

\path [draw=color3, semithick]
(axis cs:3,0.728065325689539)
--(axis cs:3,0.749347303437964);

\path [draw=color3, semithick]
(axis cs:4,0.681559893601421)
--(axis cs:4,0.707085371061773);

\path [draw=color3, semithick]
(axis cs:5,0.608418452369179)
--(axis cs:5,0.637946810277245);

\path [draw=color4, semithick]
(axis cs:0,0.874171311164846)
--(axis cs:0,0.881768708919535);

\path [draw=color4, semithick]
(axis cs:1,0.810033891862024)
--(axis cs:1,0.820542948337898);

\path [draw=color4, semithick]
(axis cs:2,0.765827174450402)
--(axis cs:2,0.782051774187911);

\path [draw=color4, semithick]
(axis cs:3,0.728938645836598)
--(axis cs:3,0.749595042507856);

\path [draw=color4, semithick]
(axis cs:4,0.680064827557026)
--(axis cs:4,0.705894115057078);

\path [draw=color4, semithick]
(axis cs:5,0.604092580583797)
--(axis cs:5,0.634172679215407);

\path [draw=color5, semithick]
(axis cs:0,0.882397404261073)
--(axis cs:0,0.898602590016881);

\path [draw=color5, semithick]
(axis cs:1,0.822639367130076)
--(axis cs:1,0.844005896868056);

\path [draw=color5, semithick]
(axis cs:2,0.785381647253338)
--(axis cs:2,0.809445726778331);

\path [draw=color5, semithick]
(axis cs:3,0.754883413220975)
--(axis cs:3,0.78042816177768);

\path [draw=color5, semithick]
(axis cs:4,0.712262586940549)
--(axis cs:4,0.73845214678555);

\path [draw=color5, semithick]
(axis cs:5,0.644412122140381)
--(axis cs:5,0.672097351967712);

%\addlegendentry{gn_no_correction}
\addplot [semithick, color0, mark=o, mark size=2, mark options={solid},dashed]
table {%
0 0.916700005531311
1 0.835449705869831
2 0.770204092675482
3 0.707177191972733
4 0.632522220325749
5 0.514647368537752
};

\addplot [semithick, color5, mark=o, mark size=2, mark options={solid},dashed]
table {%
0 0.890499997138977
1 0.833322631999066
2 0.797413687015835
3 0.767655787499328
4 0.72535736686305
5 0.658254737054047
};
%\addlegendentry{gn_no_correction}
\addplot [semithick, color4, mark=o, mark size=2, mark options={solid},dashed]
table {%
0 0.877970010042191
1 0.815288420099961
2 0.773939474319157
3 0.739266844172227
4 0.692979471307052
5 0.619132629899602
};
%\addlegendentry{gn_no_correction}

%\addlegendentry{gn_no_correction}
\addplot [semithick, color1, mark=*, mark size=2, mark options={solid}]
table {%
0 0.904611110687256
1 0.837867252310814
2 0.791783625793736
3 0.748967251582452
4 0.693278947943135
5 0.603312280704403
};
%\addlegendentry{gn_no_correction}
\addplot [semithick, color2, mark=*, mark size=2, mark options={solid}]
table {%
0 0.883689999580383
1 0.829358420873943
2 0.795155261378539
3 0.766619475264298
4 0.726993683607955
5 0.663455789026461
};
%\addlegendentry{gn_no_correction}
\addplot [semithick, color3, mark=*, mark size=2, mark options={solid}]
table {%
0 0.872889995574951
1 0.811799999914671
2 0.771954736270403
3 0.738706314563751
4 0.694322632331597
5 0.623182631323212
};

\addlegendentry{BN}
\addlegendentry{GN}
\addlegendentry{FRN}
\addlegendentry{BN (c)}
\addlegendentry{GN (c)}
\addlegendentry{FRN (c)}

\addplot [semithick, color0, mark=-, mark size=6, mark options={solid}, only marks]
table {%
0 0.914001860454317
1 0.831338594492065
2 0.76454012639293
3 0.699074955046524
4 0.622823893476053
5 0.504233351574395
};
%\addlegendentry{gn_no_correction}
\addplot [semithick, color0, mark=-, mark size=6, mark options={solid}, only marks]
table {%
0 0.919398150608305
1 0.839560817247597
2 0.775868058958034
3 0.715279428898941
4 0.642220547175445
5 0.52506138550111
};
%\addlegendentry{gn_no_correction}
\addplot [semithick, color1, mark=-, mark size=6, mark options={solid}, only marks]
table {%
0 0.901898823017998
1 0.834916205863073
2 0.788264029803453
3 0.74386839747276
4 0.68539682037221
5 0.592762051937106
};
%\addlegendentry{gn_no_correction}
\addplot [semithick, color1, mark=-, mark size=6, mark options={solid}, only marks]
table {%
0 0.907323398356513
1 0.840818298758555
2 0.795303221784019
3 0.754066105692145
4 0.701161075514061
5 0.613862509471701
};
%\addlegendentry{gn_no_correction}
\addplot [semithick, color2, mark=-, mark size=6, mark options={solid}, only marks]
table {%
0 0.875333635946893
1 0.818710570898417
2 0.783357105936799
3 0.754611732146325
4 0.715023054683454
5 0.651551841603456
};
%\addlegendentry{gn_no_correction}
\addplot [semithick, color2, mark=-, mark size=6, mark options={solid}, only marks]
table {%
0 0.892046363213874
1 0.840006270849469
2 0.806953416820279
3 0.778627218382271
4 0.738964312532456
5 0.675359736449466
};
%\addlegendentry{gn_no_correction}
\addplot [semithick, color3, mark=-, mark size=6, mark options={solid}, only marks]
table {%
0 0.868294253135301
1 0.805916305738573
2 0.763435243269797
3 0.728065325689539
4 0.681559893601421
5 0.608418452369179
};
%\addlegendentry{gn_no_correction}
\addplot [semithick, color3, mark=-, mark size=6, mark options={solid}, only marks]
table {%
0 0.877485738014601
1 0.817683694090769
2 0.780474229271008
3 0.749347303437964
4 0.707085371061773
5 0.637946810277245
};
%\addlegendentry{gn_no_correction}
\addplot [semithick, color4, mark=-, mark size=6, mark options={solid}, only marks]
table {%
0 0.874171311164846
1 0.810033891862024
2 0.765827174450402
3 0.728938645836598
4 0.680064827557026
5 0.604092580583797
};
%\addlegendentry{gn_no_correction}
\addplot [semithick, color4, mark=-, mark size=6, mark options={solid}, only marks]
table {%
0 0.881768708919535
1 0.820542948337898
2 0.782051774187911
3 0.749595042507856
4 0.705894115057078
5 0.634172679215407
};
%\addlegendentry{gn_no_correction}
\addplot [semithick, color5, mark=-, mark size=6, mark options={solid}, only marks]
table {%
0 0.882397404261073
1 0.822639367130076
2 0.785381647253338
3 0.754883413220975
4 0.712262586940549
5 0.644412122140381
};
%\addlegendentry{gn_no_correction}
\addplot [semithick, color5, mark=-, mark size=6, mark options={solid}, only marks]
table {%
0 0.898602590016881
1 0.844005896868056
2 0.809445726778331
3 0.78042816177768
4 0.73845214678555
5 0.672097351967712
};

%\addlegendentry{gn_no_correction}
\end{axis}

\end{tikzpicture}}
    \caption{Average accuracy for 5 image corruption severities and the clean data accuracy at 0 on corrupted CIFAR-10.}
    \label{fig:cifar_acc_average}
\end{minipage}
\begin{minipage}[h]{0.01\textwidth}
\hspace{\textwidth}
\end{minipage}
\begin{minipage}[h]{0.49\textwidth}
    \centering
		\resizebox{\textwidth}{!}{
		% This file was created by tikzplotlib v0.9.6.
\begin{tikzpicture}

\definecolor{color0}{rgb}{0.12156862745098,0.466666666666667,0.705882352941177}
\definecolor{color3}{rgb}{0.172549019607843,0.627450980392157,0.172549019607843}
\definecolor{color4}{rgb}{0.172549019607843,0.627450980392157,0.172549019607843}
\definecolor{color5}{rgb}{0.83921568627451,0.152941176470588,0.156862745098039}
\definecolor{color1}{rgb}{0.12156862745098,0.466666666666667,0.705882352941177}
\definecolor{color2}{rgb}{0.83921568627451,0.152941176470588,0.156862745098039}

\begin{axis}[height=5cm,width=8cm,
axis line style={white!80!black},
legend cell align={left},
legend style={fill opacity=0.8,font=\small, draw opacity=1, text opacity=1, at={(0.4,0.03)}, anchor=south east, draw=none},
tick align=outside,
tick pos=left,
x grid style={white!80!black},
xmajorgrids,
xlabel={Severity},
ylabel={Accuracy},
xmin=-0.25, xmax=5.25,
xtick style={color=white!15!black},
y grid style={white!80!black},
ymajorgrids,
ymin=0.77355456828272, ymax=0.926343083099999,
ytick style={color=white!15!black}
]
\path [draw=color0, semithick]
(axis cs:0,0.914001860454317)
--(axis cs:0,0.919398150608305);

\path [draw=color0, semithick]
(axis cs:1,0.884487534710589)
--(axis cs:1,0.892356900557389);

\path [draw=color0, semithick]
(axis cs:2,0.864816076851593)
--(axis cs:2,0.871850589285254);

\path [draw=color0, semithick]
(axis cs:3,0.908905065046389)
--(axis cs:3,0.91482826472656);

\path [draw=color0, semithick]
(axis cs:4,0.884148229153301)
--(axis cs:4,0.890318431134344);

\path [draw=color0, semithick]
(axis cs:5,0.840048105399539)
--(axis cs:5,0.851107447464536);

\path [draw=color1, semithick]
(axis cs:0,0.901898823017998)
--(axis cs:0,0.907323398356513);

\path [draw=color1, semithick]
(axis cs:1,0.866930167109617)
--(axis cs:1,0.875003178685061);

\path [draw=color1, semithick]
(axis cs:2,0.837796010966588)
--(axis cs:2,0.848515101543245);

\path [draw=color1, semithick]
(axis cs:3,0.898988350045125)
--(axis cs:3,0.904989430826796);

\path [draw=color1, semithick]
(axis cs:4,0.875029158939859)
--(axis cs:4,0.885081961072212);

\path [draw=color1, semithick]
(axis cs:5,0.837966201490035)
--(axis cs:5,0.854856029696726);

\path [draw=color2, semithick]
(axis cs:0,0.875333635946893)
--(axis cs:0,0.892046363213874);

\path [draw=color2, semithick]
(axis cs:1,0.845289820797214)
--(axis cs:1,0.869170194499722);

\path [draw=color2, semithick]
(axis cs:2,0.808995659473924)
--(axis cs:2,0.837004356261702);

\path [draw=color2, semithick]
(axis cs:3,0.874203318367734)
--(axis cs:3,0.890056663741336);

\path [draw=color2, semithick]
(axis cs:4,0.862156953093104)
--(axis cs:4,0.876763044122167);

\path [draw=color2, semithick]
(axis cs:5,0.844480596690897)
--(axis cs:5,0.858039392322775);

\path [draw=color3, semithick]
(axis cs:0,0.868294253135301)
--(axis cs:0,0.877485738014601);

\path [draw=color3, semithick]
(axis cs:1,0.820573653679874)
--(axis cs:1,0.832846365469906);

\path [draw=color3, semithick]
(axis cs:2,0.780499500774415)
--(axis cs:2,0.800400512672393);

\path [draw=color3, semithick]
(axis cs:3,0.874913417760191)
--(axis cs:3,0.881226587828341);

\path [draw=color3, semithick]
(axis cs:4,0.864921979192818)
--(axis cs:4,0.874218018766319);

\path [draw=color3, semithick]
(axis cs:5,0.844893574052815)
--(axis cs:5,0.855286443895336);

\path [draw=color4, semithick]
(axis cs:0,0.874171311164846)
--(axis cs:0,0.881768708919535);

\path [draw=color4, semithick]
(axis cs:1,0.828095192544003)
--(axis cs:1,0.839484815962772);

\path [draw=color4, semithick]
(axis cs:2,0.790105953444895)
--(axis cs:2,0.807774052391591);

\path [draw=color4, semithick]
(axis cs:3,0.879094043372993)
--(axis cs:3,0.885665974022027);

\path [draw=color4, semithick]
(axis cs:4,0.869924058366256)
--(axis cs:4,0.878375933718248);

\path [draw=color4, semithick]
(axis cs:5,0.850494953601246)
--(axis cs:5,0.861385026962871);

\path [draw=color5, semithick]
(axis cs:0,0.882397404261073)
--(axis cs:0,0.898602590016881);

\path [draw=color5, semithick]
(axis cs:1,0.853210676066628)
--(axis cs:1,0.875289320595512);

\path [draw=color5, semithick]
(axis cs:2,0.815824311243564)
--(axis cs:2,0.843895692361325);

\path [draw=color5, semithick]
(axis cs:3,0.881048722930625)
--(axis cs:3,0.896131269268319);

\path [draw=color5, semithick]
(axis cs:4,0.87016573472325)
--(axis cs:4,0.883654243513902);

\path [draw=color5, semithick]
(axis cs:5,0.853490223118456)
--(axis cs:5,0.865269778540937);

%\addlegendentry{gn-no-correction}
\addplot [semithick, color0, mark=o, mark size=2, mark options={solid},dashed]
table {%
0 0.916700005531311
1 0.888422217633989
2 0.868333333068424
3 0.911866664886475
4 0.887233330143823
5 0.845577776432037
};
\addplot [semithick, color5, mark=o, mark size=2, mark options={solid},dashed]
table {%
0 0.890499997138977
1 0.86424999833107
2 0.829860001802444
3 0.888589996099472
4 0.876909989118576
5 0.859380000829697
};
\addplot [semithick, color4, mark=o, mark size=2, mark options={solid},dashed]
table {%
0 0.877970010042191
1 0.833790004253387
2 0.798940002918243
3 0.88238000869751
4 0.874149996042252
5 0.855939990282059
};
%\addlegendentry{gn-no-correction}
\addplot [semithick, color1, mark=*, mark size=2, mark options={solid}]
table {%
0 0.904611110687256
1 0.870966672897339
2 0.843155556254917
3 0.901988890435961
4 0.880055560006036
5 0.84641111559338
};
%\addlegendentry{gn-no-correction}
\addplot [semithick, color2, mark=*, mark size=2, mark options={solid}]
table {%
0 0.883689999580383
1 0.857230007648468
2 0.823000007867813
3 0.882129991054535
4 0.869459998607635
5 0.851259994506836
};
%\addlegendentry{gn-no-correction}
\addplot [semithick, color3, mark=*, mark size=2, mark options={solid}]
table {%
0 0.872889995574951
1 0.82671000957489
2 0.790450006723404
3 0.878070002794266
4 0.869569998979568
5 0.850090008974075
};
%\addlegendentry{gn-no-correction}

%\addlegendentry{gn-no-correction}

\addlegendentry{BN}
\addlegendentry{GN}
\addlegendentry{FRN}
\addlegendentry{BN (c)}
\addlegendentry{GN (c)}
\addlegendentry{FRN (c)}

\addplot [semithick, color0, mark=-, mark size=6, mark options={solid}, only marks]
table {%
0 0.914001860454317
1 0.884487534710589
2 0.864816076851593
3 0.908905065046389
4 0.884148229153301
5 0.840048105399539
};
%\addlegendentry{gn-no-correction}
\addplot [semithick, color0, mark=-, mark size=6, mark options={solid}, only marks]
table {%
0 0.919398150608305
1 0.892356900557389
2 0.871850589285254
3 0.91482826472656
4 0.890318431134344
5 0.851107447464536
};
%\addlegendentry{gn-no-correction}
\addplot [semithick, color1, mark=-, mark size=6, mark options={solid}, only marks]
table {%
0 0.901898823017998
1 0.866930167109617
2 0.837796010966588
3 0.898988350045125
4 0.875029158939859
5 0.837966201490035
};
%\addlegendentry{gn-no-correction}
\addplot [semithick, color1, mark=-, mark size=6, mark options={solid}, only marks]
table {%
0 0.907323398356513
1 0.875003178685061
2 0.848515101543245
3 0.904989430826796
4 0.885081961072212
5 0.854856029696726
};
%\addlegendentry{gn-no-correction}
\addplot [semithick, color2, mark=-, mark size=6, mark options={solid}, only marks]
table {%
0 0.875333635946893
1 0.845289820797214
2 0.808995659473924
3 0.874203318367734
4 0.862156953093104
5 0.844480596690897
};
%\addlegendentry{gn-no-correction}
\addplot [semithick, color2, mark=-, mark size=6, mark options={solid}, only marks]
table {%
0 0.892046363213874
1 0.869170194499722
2 0.837004356261702
3 0.890056663741336
4 0.876763044122167
5 0.858039392322775
};
%\addlegendentry{gn-no-correction}
\addplot [semithick, color3, mark=-, mark size=6, mark options={solid}, only marks]
table {%
0 0.868294253135301
1 0.820573653679874
2 0.780499500774415
3 0.874913417760191
4 0.864921979192818
5 0.844893574052815
};
%\addlegendentry{gn-no-correction}
\addplot [semithick, color3, mark=-, mark size=6, mark options={solid}, only marks]
table {%
0 0.877485738014601
1 0.832846365469906
2 0.800400512672393
3 0.881226587828341
4 0.874218018766319
5 0.855286443895336
};
%\addlegendentry{gn-no-correction}
\addplot [semithick, color4, mark=-, mark size=6, mark options={solid}, only marks]
table {%
0 0.874171311164846
1 0.828095192544003
2 0.790105953444895
3 0.879094043372993
4 0.869924058366256
5 0.850494953601246
};
%\addlegendentry{gn-no-correction}
\addplot [semithick, color4, mark=-, mark size=6, mark options={solid}, only marks]
table {%
0 0.881768708919535
1 0.839484815962772
2 0.807774052391591
3 0.885665974022027
4 0.878375933718248
5 0.861385026962871
};
%\addlegendentry{gn-no-correction}
\addplot [semithick, color5, mark=-, mark size=6, mark options={solid}, only marks]
table {%
0 0.882397404261073
1 0.853210676066628
2 0.815824311243564
3 0.881048722930625
4 0.87016573472325
5 0.853490223118456
};
%\addlegendentry{gn-no-correction}
\addplot [semithick, color5, mark=-, mark size=6, mark options={solid}, only marks]
table {%
0 0.898602590016881
1 0.875289320595512
2 0.843895692361325
3 0.896131269268319
4 0.883654243513902
5 0.865269778540937
};

%\addlegendentry{gn-no-correction}
\end{axis}

\end{tikzpicture}}
    \caption{Accuracy for 5 saturation change severites and the clean data accuracy at 0 on corrupted CIFAR-10.}
    \label{fig:cifar_acc_saturate}
\end{minipage}
\end{figure}

\begin{figure}[h]
    \centering
		\resizebox{0.5\textwidth}{!}{
		% This file was created by tikzplotlib v0.9.4.
\begin{tikzpicture}

\definecolor{color0}{rgb}{0.12156862745098,0.466666666666667,0.705882352941177}
\definecolor{color1}{rgb}{1,0.498039215686275,0.0549019607843137}

\begin{axis}[height=5cm,width=8cm,
axis line style={white!80!black},
legend cell align={left},
legend style={fill opacity=0.8,font=\small, draw opacity=1, text opacity=1, draw=none},
tick align=outside,
tick pos=left,
x grid style={white!80!black},
xmajorgrids,
xmin=-0.25, xmax=5.25,
xlabel={Severity},
ylabel={Accuracy},
xtick style={color=white!15!black},
y grid style={white!80!black},
ymajorgrids,
ymin=0.565607895898191, ymax=0.916971076554374,
ytick style={color=white!15!black}
]
\path [draw=color0, semithick]
(axis cs:0,0.899900019168854)
--(axis cs:0,0.899900019168854);

\path [draw=color0, semithick]
(axis cs:1,0.829878951373853)
--(axis cs:1,0.829878951373853);

\path [draw=color0, semithick]
(axis cs:2,0.783384210185001)
--(axis cs:2,0.783384210185001);

\path [draw=color0, semithick]
(axis cs:3,0.738147368556575)
--(axis cs:3,0.738147368556575);

\path [draw=color0, semithick]
(axis cs:4,0.678110530501918)
--(axis cs:4,0.678110530501918);

\path [draw=color0, semithick]
(axis cs:5,0.581578949564382)
--(axis cs:5,0.581578949564382);

\path [draw=color1, semithick]
(axis cs:0,0.901000022888184)
--(axis cs:0,0.901000022888184);

\path [draw=color1, semithick]
(axis cs:1,0.840047362603639)
--(axis cs:1,0.840047362603639);

\path [draw=color1, semithick]
(axis cs:2,0.795668423175812)
--(axis cs:2,0.795668423175812);

\path [draw=color1, semithick]
(axis cs:3,0.751615787807264)
--(axis cs:3,0.751615787807264);

\path [draw=color1, semithick]
(axis cs:4,0.693294729057111)
--(axis cs:4,0.693294729057111);

\path [draw=color1, semithick]
(axis cs:5,0.599642112066871)
--(axis cs:5,0.599642112066871);
\addplot [semithick, color0, mark=*, mark size=2, mark options={solid}]
table {%
0 0.899900019168854
1 0.829878951373853
2 0.783384210185001
3 0.738147368556575
4 0.678110530501918
5 0.581578949564382
};
%\addlegendentry{bn-lamb-1.0-0.25-severity}
\addplot [semithick, color1, mark=*, mark size=2, mark options={solid}]
table {%
0 0.901000022888184
1 0.840047362603639
2 0.795668423175812
3 0.751615787807264
4 0.693294729057111
5 0.599642112066871
};
\addlegendentry{BN (c) without sparsity term}
\addlegendentry{BN (c) with sparsity term}
\addplot [semithick, color0, mark=+, mark size=5, mark options={solid}, only marks]
table {%
0 0.899900019168854
1 0.829878951373853
2 0.783384210185001
3 0.738147368556575
4 0.678110530501918
5 0.581578949564382
};
%\addlegendentry{bn-lamb-1.0-0.25-severity}
\addplot [semithick, color0, mark=+, mark size=5, mark options={solid}, only marks]
table {%
0 0.899900019168854
1 0.829878951373853
2 0.783384210185001
3 0.738147368556575
4 0.678110530501918
5 0.581578949564382
};
%\addlegendentry{bn-lamb-1.0-0.25-severity}
\addplot [semithick, color1, mark=+, mark size=5, mark options={solid}, only marks]
table {%
0 0.901000022888184
1 0.840047362603639
2 0.795668423175812
3 0.751615787807264
4 0.693294729057111
5 0.599642112066871
};
%\addlegendentry{bn-lamb-1.0-0.25-severity}
\addplot [semithick, color1, mark=+, mark size=5, mark options={solid}, only marks]
table {%
0 0.901000022888184
1 0.840047362603639
2 0.795668423175812
3 0.751615787807264
4 0.693294729057111
5 0.599642112066871
};
%\addlegendentry{bn-lamb-1.0-0.25-severity}

\end{axis}

\end{tikzpicture}}
    \caption{Average accuracy for image corruption severities between a BN model with/without a sparsity term in the correction.}
    \label{fig:sparsity}
\end{figure}
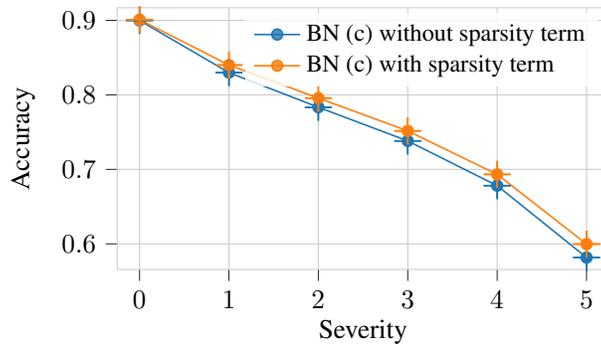

The results in Table \ref{tab:cifar10} again show the average classification accuracy of the randomly initialized models over all corruption severities. Similar as for the MNIST datasets, the models using BN show the most improvement over all datasets, increasing the accuracy by $3.38~\%$ for the corrected networks. For the other normalizations, i.e GN and FRN, we do not see a large performance difference. However, this comes from the fact that for low-intensity corruptions the correction method achieves slightly lower classification accuracy, whereas the performance increases for high-severity corruptions. This behavior can be seen in Figure \ref{fig:cifar_acc_average} for the average accuracy over the corruption severity. Detailed results for the highest severity are listed in Appendix \ref{tab:cifar10_severity_5}. Figure \ref{fig:cifar_acc_saturate} shows the performance change with the corruption severity for the special case of saturation corruptions. We see that at low levels our method decreases the performance, but can even improve performance as the corruption level increases. In Figure \ref{fig:sparsity}, we see the performance difference with and without the sparsity term in the prior of the correction approach using BN (see~\eqref{eq:prior_term}).  

\subsection{Corrupted ImageNet classification}\label{sec:res_imagenet}
The corrupted ILSVRC 2012 dataset contains 19 different corruption variants of the original dataset. It contains 50000 RGB images of size 224$\times$224 per corruption that we used for our evaluations. 
For the evaluation on the corrupted ImageNet (ILSVRC 2012) dataset we again choose the same parameter set $\lambda_1=0.5$ and $\lambda_2=0.5$ and $N_{iter}=1$ for all normalization methods. We trained a single ResNet-50 model for 90 epochs using an SGD optimizer with a base learning rate of 0.1 on the clean ImageNet dataset using an implementation adapted from Tensorflow Model Garden~\cite{ILSVRC15}. The learning rate was decayed after 30, 60, and 80 epochs, and a warm-up from 0.02 to 0.1 was used during the first five epochs of training.

\begin{table}[H]
    \centering
        \caption{Classification performance in $\%$ on the ImageNet-C dataset over all 5 severities. (c) indicates the model with the distribution correction enabled. The mean Corruption Error (mCE) is calculated using a AlexNet baseline~\cite{hendrycks2018benchmarking}.}
    \tiny
    \begin{tabular}{c||c|c||c|c||c|c}
         noise type&BN&BN (c)&FRN& FRN (c)&GN&GN (c)\\
         \hline\hline
         %identity&76.64~\%&&76.60~\%&75.99~\%&\textbf{76.76}~\%\\
         brightness      &68.06~\%&67.13~\%&66.53~\%&66.51~\%&\textbf{67.83}~\%&67.62~\%\\\hline
         contrast        &40.39~\%&42.68~\%&39.39~\%&44.82~\%&56.93~\%&\textbf{58.28}~\%\\\hline
         defocus blur    &36.90~\%&36.55~\%&35.92~\%&36.60~\%&36.50~\%&\textbf{37.80}~\%\\\hline
         elastic         &45.16~\%&46.01~\%&46.06~\%&45.90~\%&47.88~\%&\textbf{48.26}~\%\\\hline
         fog             &56.83~\%&58.01~\%&52.46~\%&54.97~\%&61.34~\%&\textbf{61.61}~\%\\\hline
         frost           &41.05~\%&43.33~\%&40.34~\%&42.80~\%&42.01~\%&\textbf{44.28}~\%\\\hline
         gaussian blur   &40.51~\%&40.03~\%&39.16~\%&39.71~\%&39.62~\%&\textbf{41.12}~\%\\\hline
         gaussian noise  &35.96~\%&38.44~\%&34.52~\%&36.11~\%&40.48~\%&\textbf{41.92}~\%\\\hline
         glass blur      &25.93~\%&26.61~\%&27.98~\%&27.86~\%&28.87~\%&\textbf{29.19}~\%\\\hline
         impulse noise   &31.32~\%&34.16~\%&31.30~\%&33.28~\%&37.06~\%&\textbf{38.68}~\%\\\hline
         jpeg compression&\textbf{57.36}~\%&56.35~\%&52.24~\%&52.25~\%&56.25~\%&56.28~\%\\\hline
         motion blur     &34.75~\%&36.24~\%&39.39~\%&39.90~\%&40.56~\%&\textbf{40.88}~\%\\\hline
         pixelate        &62.06~\%&61.99~\%&59.59~\%&59.80~\%&61.87~\%&\textbf{63.62}~\%\\\hline
         saturate        &62.81~\%&62.69~\%&62.22~\%&61.68~\%&\textbf{63.39}~\%&63.14~\%\\\hline
         shot noise      &33.69~\%&35.94~\%&32.99~\%&34.59~\%&38.05~\%&\textbf{39.39}~\%\\\hline
         snow            &35.52~\%&36.83~\%&40.55~\%&41.82~\%&41.65~\%&\textbf{43.38}~\%\\\hline
         spatter         &53.34~\%&53.62~\%&53.52~\%&54.28~\%&55.84~\%&\textbf{56.46}~\%\\\hline
         speckle noise   &41.52~\%&42.91~\%&41.59~\%&42.78~\%&45.39~\%&\textbf{46.41}~\%\\\hline
         zoom blur       &37.40~\%&\textbf{38.24}~\%&34.98~\%&35.53~\%&36.62~\%&37.46~\%\\\hline
         \hline
         \textbf{avg. top 1 accuracy}         &44.24~\%&45.15~\%&43.72~\%&44.80~\%&47.27~\%&\textbf{48.20}~\%\\
         \hline
         \textbf{mCE}&70.95~\%&69.93~\%&71.75~\%&70.49~\%&67.35~\%&\textbf{66.23}~\%\\
\end{tabular}
\vspace{-0.5cm}
\end{table}\label{tab:imagenet_results}
The results in Table \ref{tab:imagenet_results} show that for the corrupted ImageNet dataset, all models regardless of the used norm achieve about 1\% performance improvement by using the proposed correction method. Generally, we see that the more flexible GN models are the most robust against image corruptions outperforming BN and FRN by $>$3.0\%. This supports our assumption that even for large amounts of data, covariate shift still influences performance for corrupted inputs. Also, we see that unlike for CIFAR-10, here also FRN and GN improve for corruptions with lower severity. This might be due to the more complicated distribution shapes of ImageNet activations.
\subsection{Limitations}\label{sec:limitations}
The main limitation of this approach is the additional computational complexity. As the algorithm requires the sorting of all $N$ activations of a layer, the complexity scales with $\mathcal{O}(N\log{}N)$. This causes an overhead, especially for datasets with large images, such as ImageNet, and also limits the number of possible iterations which can be used to converge the algorithm. For MNIST the average evaluation time of a single sample increases from 0.3 ms to 2 ms (1 GPU), whereas on ImageNet the average evaluation time rises from 0.1 ms to 12 ms per image (4 GPUs). Furthermore, as we can only use the layer-wise distribution as a target, we introduce distortions to the individual channels causing performance degradation for clean data and some low-intensity corruptions.

%We do not have particular ethical concerns regarding our proposed method and do not expect a negative societal impact. One should note, however, that any method that makes image classifiers more robust inevitably broadens the field of potential applications. While we are positive that the research community as whole values and pursues social good, one cannot neglect that computer vision has been put to questionable use in the past. Along those lines, we are aware that research on computer vision has traditionally been of great interest for the military and for automated decision processes that might make unfair or unethical decisions for minorities not sufficiently represented in the data.
\section{Conclusion and Outlook}\label{sec:conclusion}
We proposed a non-parametric activation distribution correction method based on the Wasserstein distance. It reduces the mismatch between test-time and training distributions of the activations within DNNs. The proposed method uses a maximum a-posteriori estimate, determined by minimizing the energy with respect to a data likelihood term and a prior term based on the Wasserstein distance. Our proposed method works in an unsupervised setting and can be retrofitted into existing networks without retraining. In our experiments, we showed that our correction algorithm can effectively reduce the mismatch between test-time and training distributions. This results in improved classification performance on corrupted input data, as we have evaluated for the corrupted variants of MNIST, CIFAR-10, and ImageNet (ILSVRC 2012). The results show that the proposed method is particularly effective for strong input corruptions and increases overall robustness for most of the investigated models.
For future applications, we want to further evaluate the capabilities of this method regarding robustness and also explore the use of our algorithm for reducing the impact of parametric approximations.
\FloatBarrier
\bibliographystyle{unsrt}
\bibliography{main.bbl}%{refs.bib}

\appendix
\section{Analyzing the distributions}\label{sec:target_dist}
Since we assume a quasi-stationary distribution of the activations of the flattened layer across the training set, we need to verify if this assumption holds for our ResNet models. This is necessary in order to create a meaningful target distribution $q(\mathbf{t})$. Therefore, we analyze the variance $\sigma^2$ of the target values $\mathbf{t}$ over the training set:
\begin{equation}
    \sigma^2_{(i)}=\frac{1}{M}\sum_{m=1}^M (t_{(i)}-a^{(m)}_{(i)})^2.
\end{equation}
\begin{figure}[h]
    \centering
	\begin{minipage}[h]{0.32\textwidth}
		\includegraphics[width=\textwidth]{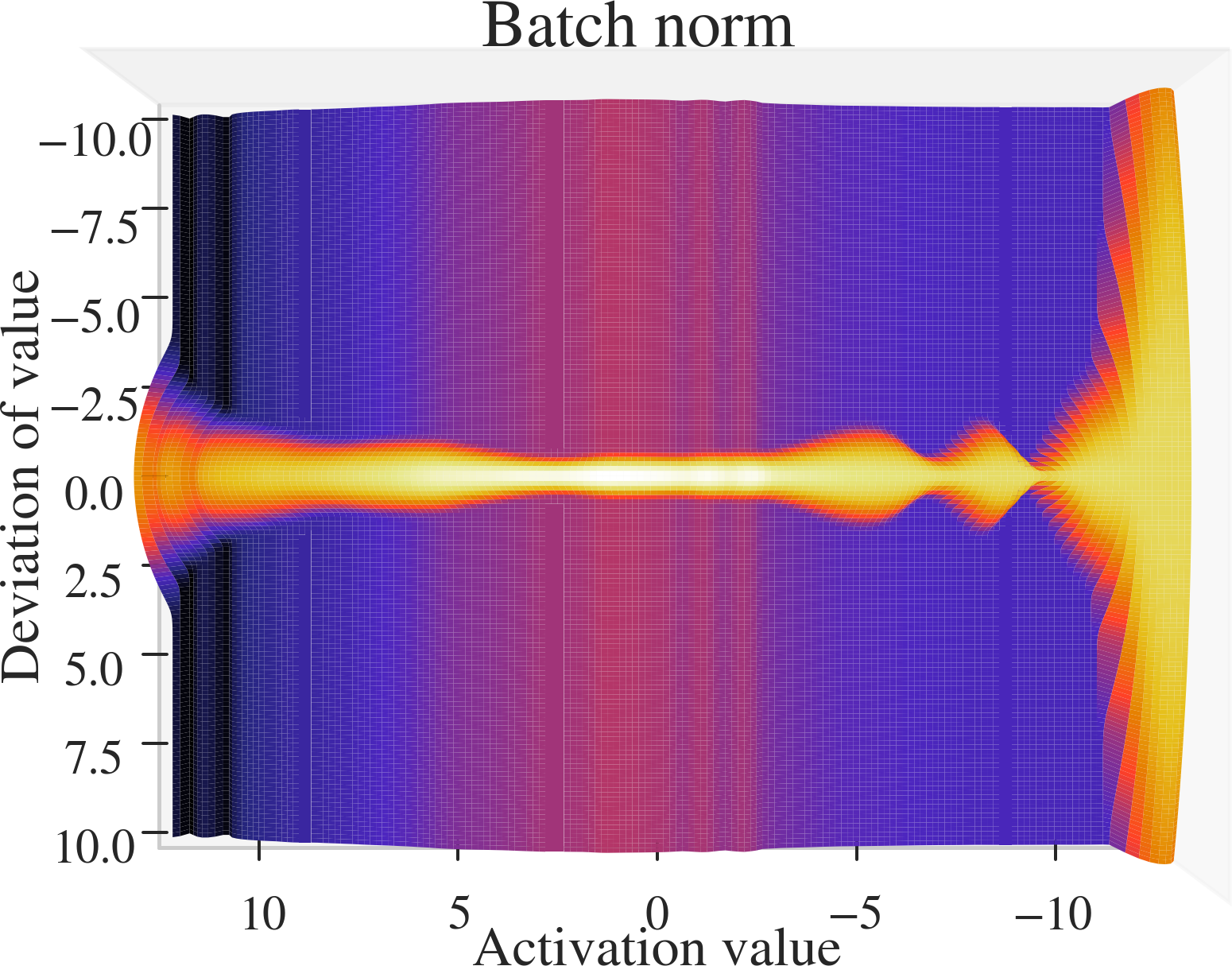}
    \end{minipage}
    	\begin{minipage}[h]{0.32\textwidth}
		\includegraphics[width=\textwidth]{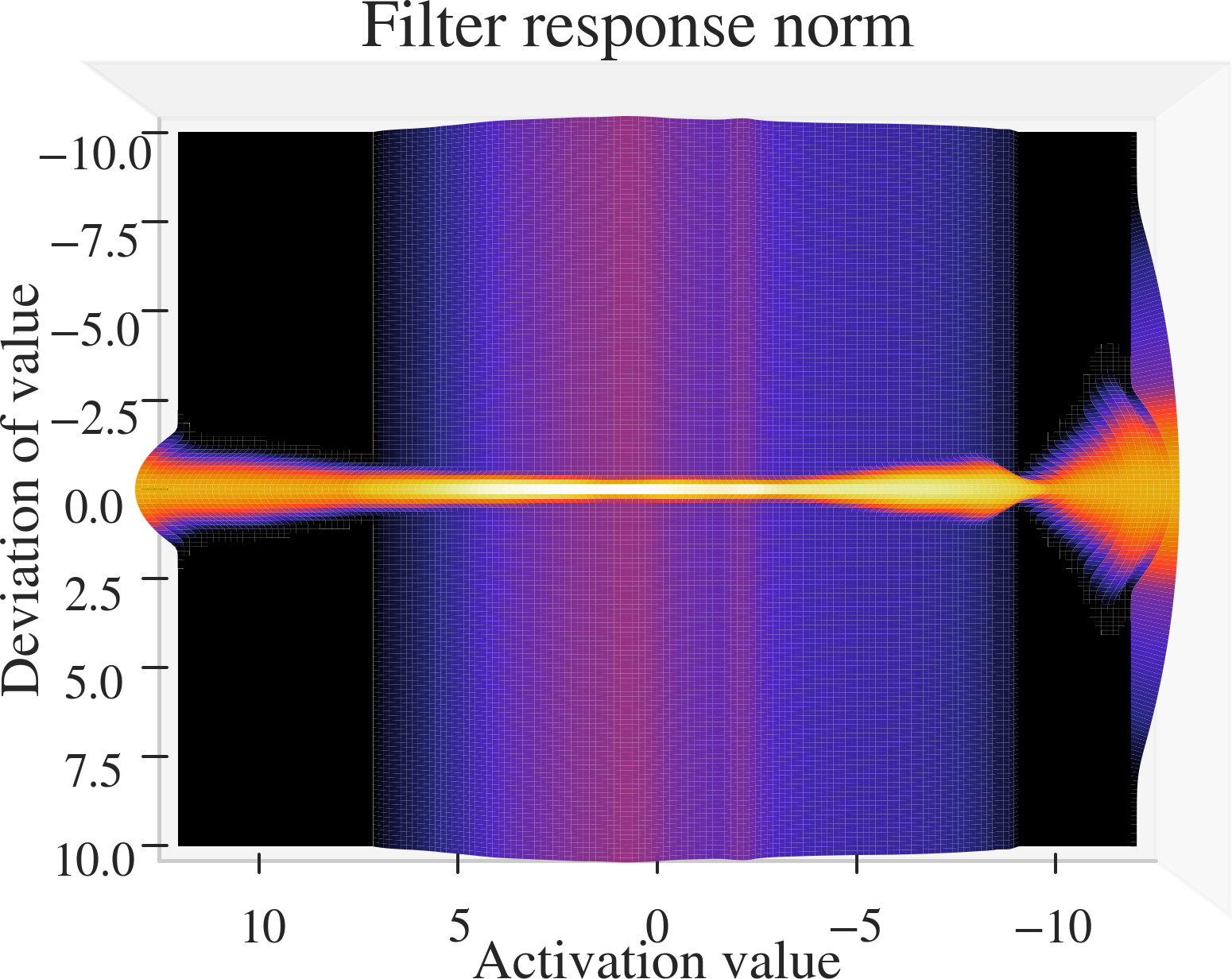}
    \end{minipage}
    \begin{minipage}[h]{0.32\textwidth}
		\includegraphics[width=\textwidth]{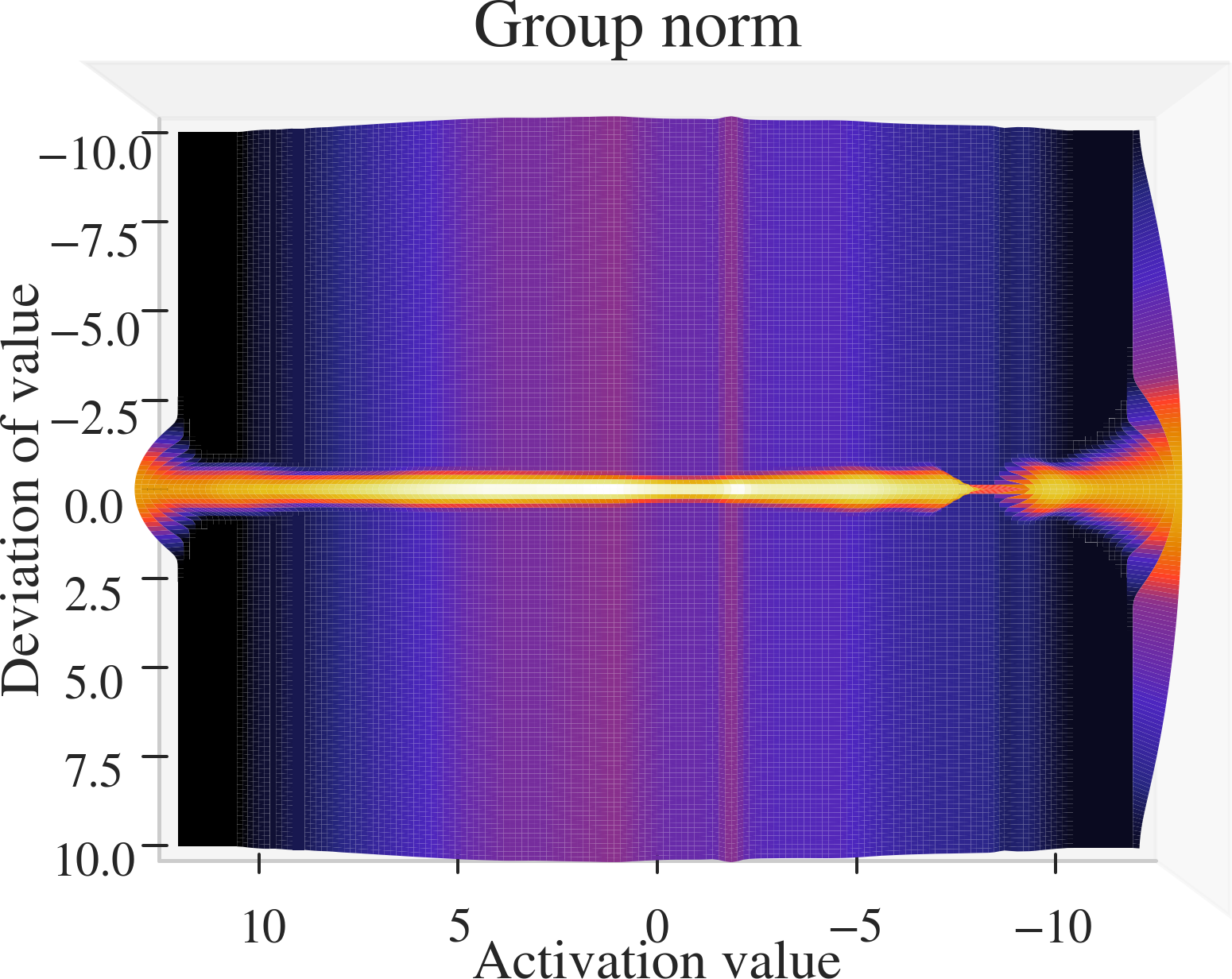}
    \end{minipage}
    \caption{The log probability distributions $\log(p(x_{(i)}|t_{(i)},\sigma^2(t_{(i)})))$ of a sorted value $(x_{(i)})$ given the corresponding target values $t_{(i)}$ and variances $\sigma^2(t_{(i)})$, assuming Gaussian distributed target values. Here $\log(p(x_{(i)}|t_{(i)},\sigma^2(t_{(i)})))$ is shown for the different normalization methods BN, FRN and GN for activations before the first ReLU function. The x-axis shows the mean value of the $i^{th}$ sorted target value and the y-axis shows the deviation from the mean over the training set.}    \label{fig:mean_var}
\end{figure}
\begin{figure}
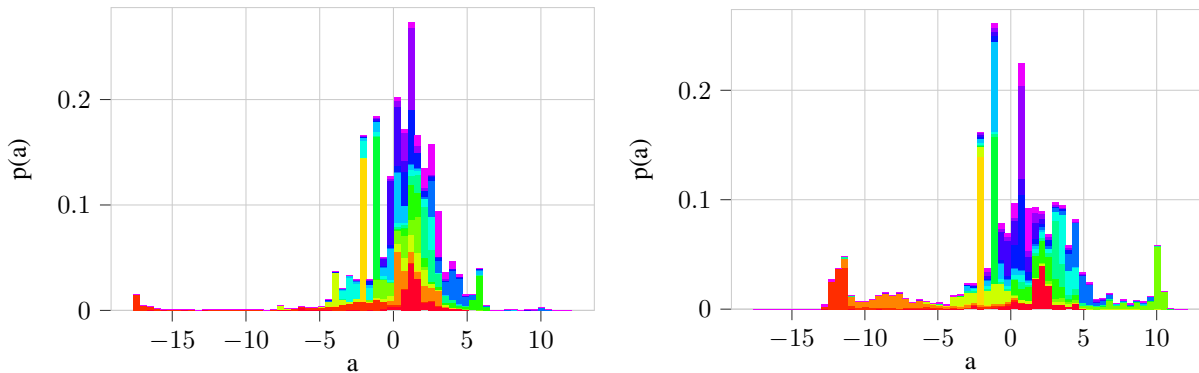

    \centering
	\begin{minipage}[h]{0.49\textwidth}
				\resizebox{\textwidth}{!}{
		\input{orig_sample_0.tikz}}
    \end{minipage}
    \begin{minipage}[h]{0.49\textwidth}
		\resizebox{\textwidth}{!}{
		\input{max_dist_sample_to_0.tikz}}
    \end{minipage}
    \caption{The distribution over all channels of the first layer of a ResNet-20 classifier trained on CIFAR-10 before the ReLU activation. Both images shows the distribution of activations over all channels stacked, indicated by using a different color for each channel. The left image is the test sample $k$ for which the image with the largest Wasserstein distance within the entire test set (with respect to channel distributions) was selected. The most dissimilar image $\tilde{m}$ is shown on the right.}\label{fig:max_diff}
\end{figure}
In Figure \ref{fig:mean_var}, we see that although all normalization methods have a similar shape for the variance of their target value distribution, BN experiences a much higher variation at the tails of the distributions, having a value range 5 times higher than GN and FRN. This observation is in line with the intuition that more flexible normalization methods produce narrower activation distributions. 

Generally, we see that with the exception of the distribution-tails, the distribution of the targets is more or less stationary. This is an interesting observation, as one would expect that the activation distributions for the entire layer have a higher dependency on the specific input.
Following this observation, we analyze this behavior by comparing the activation distributions for the most dissimilar examples in the test set.
Therefore we calculate the Wasserstein distances from a test example $k$ to all other test examples $m$ with respect to their channel distributions, utilizing the order statistics for each channel $j$ separately and select the sample with the largest total value with index $\tilde{m}$,
\begin{equation}
   \tilde{m} = \argmax_m \sum_{i,j}|a_{j,(i)}^{(k)}-a_{j,(i)}^{(m)}|.
\end{equation}

Figure \ref{fig:max_diff} shows that while the individual channel distributions are very distinct, the shape of the distribution over all channels is more or less stable. This indicates that the inverse correlation of channels can conserve the overall shape of the distribution for each layer.
\section{Algorithm}\label{sec:algo}
The algorithm is described in the following pseudo-code. The algorithm is executed for every layer and returns the corrected activations $\tilde{\mathbf{a}}$. The step-sizes $\lambda_1$ and $\lambda_2$ determine the relative importance of the prior term versus the likelihood term.

\begin{algorithm}
\caption{Activation Correction Algorithm}
\hspace*{\algorithmicindent} \textbf{Input:} Unsorted activations $\mathbf{a}$, sorted target values $\mathbf{t}$, step-sizes $\lambda_1, \lambda_2$, and maximum number of iterations $N_{iter}$  \\
\hspace*{\algorithmicindent} \textbf{Output:} Corrected activations $\tilde{\mathbf{a}}$ 
\begin{algorithmic}
\State $N \gets \mathrm{len}(\mathbf{a})$
\State $\tilde{\mathbf{a}} \gets \mathbf{a}-\frac{1}{N} \sum_i a_i$
\For{$n < N_{iter}$}
    \State [$a_{(i)}],\mathbf{j} \gets \mathrm{sort}(\tilde{\mathbf{a}})$
    \For{$i< N$}
        \State $j \gets j_i$
        \State $\tilde{a}_j \gets \tilde{a}_j + \lambda_1 \cdot (t_{(i)}-a_{(i)})$
        \State $\tilde{a}_j  \gets \tilde{a}_j + \lambda_2 \cdot
            (a_j-\tilde{a}_j)$
    \EndFor

\EndFor
    \For{$j< N$}
        \If{$a_j \ne 0$}
            \State $\tilde{a}_j \gets \tilde{a}_j$
        \Else
            \State $\tilde{a}_j \gets a_j$
        \EndIf
    \EndFor
\end{algorithmic}\label{alg:algo}
\end{algorithm}

\section{Supplemental material of experiments}\label{sec:imagenet_sev5}
\begin{table}[h]
    \centering

        \caption{Step size parameters and number of iterations for the correction algorithm on the MNIST dataset.}
    \tiny
    \begin{tabular}{c|c|c|c}
         Norm& $\lambda_1$&$\lambda_2$&$N_{iter}$ \\
         \hline
         BN&0.75&0.25&2\\
         FRN&0.25&0.5&1\\
         GN&0.5&0.5&1\\
    \end{tabular}
    \label{tab:mnist_set}
\end{table}
\begin{table}[h]
    \centering
        \caption{Classification accuracy in $\%$ on the corrupted CIFAR-10 dataset with corruption severity 5 averaged over 10 randomly initialized models. (c) indicates the model with the distribution correction enabled.}
    \tiny
    \begin{tabular}{c||c|c||c|c||c|c}
         noise type&BN&BN (c)&FRN& FRN (c)&GN&GN (c)\\
         \hline
 brightness&78.84$\pm$22.28&78.62$\pm$21.96&83.99$\pm$0.64&83.44$\pm$0.6&\textbf{85.16$\pm$0.89}&84.74$\pm$0.95 \\ \hline
 contrast&25.77$\pm$6.49&60.08$\pm$16.88&74.33$\pm$1.84&75.21$\pm$1.82&85.18$\pm$1.33&\textbf{85.54$\pm$1.13} \\ \hline
 defocus blur&44.72$\pm$11.7&64.42$\pm$17.92&67.14$\pm$3.58&67.54$\pm$2.99&\textbf{73.95$\pm$1.72}&73.82$\pm$1.41 \\ \hline
 elastic&62.14$\pm$17.28&60.3$\pm$16.42&65.96$\pm$1.46&65.2$\pm$1.48&\textbf{70.35$\pm$0.87}&69.21$\pm$0.98 \\ \hline
 fog&57.54$\pm$15.62&62.97$\pm$17.61&71.38$\pm$2.06&71.23$\pm$1.94&\textbf{72.6$\pm$1.59}&72.29$\pm$1.45 \\ \hline
 frost&50.15$\pm$12.5&57.32$\pm$15.12&64.19$\pm$2.37&64.57$\pm$2.21&69.59$\pm$2.55&\textbf{70.5$\pm$2.42} \\ \hline
 frosted glass blur&39.5$\pm$10.03&40.87$\pm$10.26&45.27$\pm$2.14&45.39$\pm$2.1&\textbf{49.51$\pm$2.76}&49.48$\pm$2.32 \\ \hline
 gaussian blur&29.75$\pm$7.01&49.48$\pm$14.08&53.36$\pm$4.84&54.08$\pm$4.72&64.42$\pm$3.26&\textbf{65.1$\pm$2.69} \\ \hline
 gaussian noise&17.69$\pm$3.22&25.82$\pm$6.6&34.32$\pm$5.96&37.16$\pm$5.42&32.93$\pm$4.16&\textbf{37.63$\pm$3.2} \\ \hline
 impulse noise&25.97$\pm$6.48&34.88$\pm$9.23&43.01$\pm$4.28&44.36$\pm$3.83&44.89$\pm$3.82&\textbf{46.11$\pm$2.51} \\ \hline
 jpeg compression&62.45$\pm$17.41&58.32$\pm$15.75&63.26$\pm$0.71&62.53$\pm$0.59&\textbf{68.42$\pm$1.34}&66.84$\pm$1.39 \\ \hline
 motion blur&51.55$\pm$13.82&62.13$\pm$17.09&70.39$\pm$1.85&70.37$\pm$1.76&\textbf{77.25$\pm$1.35}&76.78$\pm$1.28 \\ \hline
 pixelate&33.94$\pm$8.15&36.49$\pm$8.93&43.88$\pm$2.8&43.59$\pm$3.04&49.65$\pm$5.22&\textbf{49.73$\pm$5.3} \\ \hline
 saturate&77.04$\pm$22.54&77.25$\pm$22.2&85.59$\pm$0.54&85.01$\pm$0.52&\textbf{85.94$\pm$0.59}&85.13$\pm$0.68 \\ \hline
 shot noise&33.23$\pm$8.28&43.8$\pm$11.54&52.1$\pm$4.96&54.31$\pm$4.39&51.13$\pm$4.31&\textbf{54.53$\pm$3.03} \\ \hline
 snow&60.76$\pm$16.62&63.4$\pm$16.91&68.64$\pm$1.3&68.44$\pm$1.35&73.11$\pm$1.88&\textbf{73.34$\pm$1.85} \\ \hline
 spatter&63.37$\pm$17.51&68.71$\pm$19.15&73.32$\pm$1.01&73.6$\pm$1.0&73.39$\pm$1.18&\textbf{74.43$\pm$1.15} \\ \hline
 speckle noise&37.19$\pm$9.3&45.26$\pm$11.66&53.29$\pm$4.21&\textbf{54.88$\pm$3.78}&51.64$\pm$3.77&54.27$\pm$2.68 \\ \hline
 zoom blur&48.51$\pm$12.84&63.02$\pm$17.36&62.96$\pm$2.86&63.13$\pm$2.54&\textbf{71.58$\pm$1.21}&71.09$\pm$1.39 \\ \hline
 identity&83.52$\pm$24.44&82.51$\pm$23.85&87.8$\pm$0.38&87.29$\pm$0.46&\textbf{89.05$\pm$0.81}&88.37$\pm$0.84 \\ \hline\hline
 \textbf{average}&49.18$\pm$12.91&56.78$\pm$15.19&63.21$\pm$1.43&63.57$\pm$1.41&66.99$\pm$1.33&\textbf{67.45$\pm$1.15} \\ 
\end{tabular}
\end{table}\label{tab:cifar10_severity_5}
\begin{table}[h]
    \centering
        \caption{Classification accuracy in $\%$ on the corrupted ImageNet dataset with corruption severity 5. (c) indicates the model with the distribution correction enabled.}
    \tiny
    \begin{tabular}{c||c|c||c|c||c|c}
         noise type&BN&BN (c)&FRN& FRN (c)&GN&GN (c)\\
         \hline
         %identity&76.64~\%&&76.60~\%&75.99~\%&\textbf{76.76}~\%\\
         brightness&58.15~\%&56.60~\%&57.71~\%&58.11~\%&\textbf{58.32}~\%&58.14~\%\\\hline
         contrast&5.38~\%&8.63~\%&5.71~\%&12.24~\%&26.39~\%&\textbf{30.12}~\%\\\hline
         defocus blur&15.70~\%&14.49~\%&14.57~\%&15.95~\%&14.43~\%&\textbf{15.63}~\%\\\hline
         elastic&14.90~\%&16.85~\%&18.31~\%&18.58~\%&20.00~\%&\textbf{20.55}~\%\\\hline
         fog&44.66~\%&47.85~\%&39.35~\%&43.25~\%&49.99~\%&\textbf{50.94}~\%\\\hline
         frost&26.48~\%&29.99~\%&26.45~\%&30.09~\%&28.09~\%&\textbf{31.28}\%\\\hline
         gaussian blur&\textbf{12.97}~\%&10.99~\%&11.68~\%&12.83~\%&11.50~\%&12.30~\%\\\hline
         gaussian noise&4.04~\%&7.19~\%&4.48~\%&5.71~\%&9.74~\%&\textbf{10.98}~\%\\\hline
         glass blur &7.78~\%&9.01~\%&9.94~\%&\textbf{10.07}~\%&8.54~\%&9.02~\%\\\hline
         impulse noise&4.36~\%&8.07~\%&5.19~\%&6.79~\%&10.70~\%&\textbf{12.05}~\%\\\hline
         jpeg compression&\textbf{42.73}~\%&41.5~\%&30.64~\%&31.22~\%&39.58~\%&39.14~\%\\\hline
         motion blur&9.29~\%&11.20~\%&13.83~\%&14.95~\%&14.27~\%&\textbf{14.99}~\%\\\hline
         pixelate&51.04~\%&50.72~\%&49.28~\%&50.05~\%&51.57~\%&\textbf{54.26}~\%\\\hline
         saturate&49.22~\%&50.62~\%&51.48~\%&51.35~\%&\textbf{53.18}~\%&52.97~\%\\\hline
         shot noise&5.57~\%&8.59~\%&6.27~\%&7.84~\%&10.86~\%&\textbf{12.08}~\%\\\hline
         snow&20.65~\%&22.26~\%&26.98~\%&28.89~\%&27.19~\%&\textbf{29.38}~\%\\\hline
         spatter&30.49~\%&32.38~\%&32.64~\%&34.56~\%&35.27~\%&\textbf{36.38}~\%\\\hline
         speckle noise&16.01~\%&19.56~\%&19.98~\%&18.06~\%&22.63~\%&\textbf{23.80}~\%\\\hline
         zoom blur&24.14~\%&\textbf{25.10}~\%&22.68~\%&23.78~\%&23.94~\%&24.85~\%\\\hline
         \hline
         \textbf{average}&23.53~\%&24.82~\%&23.43~\%&25.06~\%&27.17~\%&\textbf{28.38}~\%\\
\end{tabular}
\end{table}\label{tab:imagenet_severity_5}
\end{document}